\documentclass{article}
\PassOptionsToPackage{table}{xcolor}
\usepackage{times}

    \usepackage[final]{neurips_2024}

\usepackage{amsmath,amsfonts,bm}

\def\eqref#1{equation~\ref{#1}}

\def\1{\bm{1}}

\DeclareMathAlphabet{\mathsfit}{\encodingdefault}{\sfdefault}{m}{sl}
\SetMathAlphabet{\mathsfit}{bold}{\encodingdefault}{\sfdefault}{bx}{n}

\usepackage{floatflt}
\usepackage{hyperref}
\usepackage{url}
\usepackage{graphicx}
\usepackage{booktabs} 
\usepackage{enumitem}
\usepackage{pifont}
\usepackage{multirow}   
\usepackage{subcaption}
\usepackage{wrapfig}
\usepackage{bbm}
\usepackage[table]{xcolor}
\usepackage{xspace}
\usepackage[most]{tcolorbox}
\usepackage{threeparttable} 
\usepackage{siunitx} 
\usepackage{tabularx}
\usepackage{makecell}
\usepackage{float}
\usepackage{placeins}
\usepackage{algorithm}
\usepackage{algpseudocode}

\newcommand{\github}{\raisebox{-1.5pt}{\includegraphics[height=1.05em]{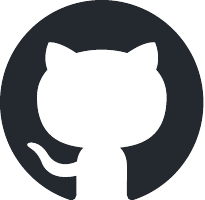}}\xspace}
\newcommand{\huggingface}{\raisebox{-1.5pt}{\includegraphics[height=1.05em]{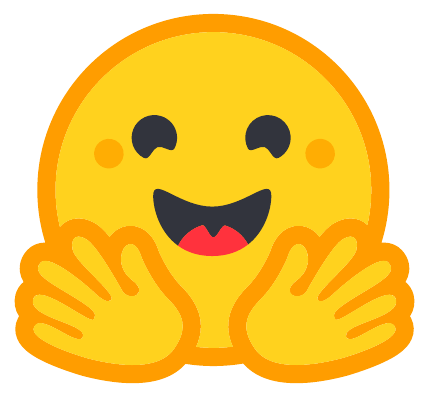}}\xspace}

\newcommand{\hhide}[1]{}
\newcommand{\hide}[1]{}
\newcommand{\tablehead}[1]{\textbf{#1}}

\definecolor{takeaway}{RGB}{165, 209, 216}
\definecolor{takeawayTitle}{RGB}{57, 89, 163}

\title{MiniCPM-V 4.5: Cooking Efficient MLLMs via Architecture, Data, and Training Recipes}

\author{\textbf{Tianyu Yu} \quad \textbf{Zefan Wang} \quad \textbf{Chongyi Wang} \quad \textbf{Fuwei Huang} \quad \textbf{Wenshuo Ma} \quad \textbf{Zhihui He} \\
\textbf{Tianchi Cai} \quad \textbf{Weize Chen} \quad \textbf{Yuxiang Huang} \quad \textbf{Yuanqian Zhao} \quad \textbf{Bokai Xu} \quad \textbf{Junbo Cui} \\
\textbf{Yingjing Xu} \quad \textbf{Liqing Ruan} \quad \textbf{Luoyuan Zhang} \quad \textbf{Hanyu Liu} \quad \textbf{Jingkun Tang} \quad \textbf{Hongyuan Liu} \\ \textbf{Qining Guo} \quad
\textbf{Wenhao Hu} \quad \textbf{Bingxiang He} \quad \textbf{Jie Zhou} \quad \textbf{Jie Cai}  \quad \textbf{Ji Qi} \quad \textbf{Zonghao Guo} \\ \textbf{Chi Chen} \quad \textbf{Guoyang Zeng} \quad
\textbf{Yuxuan Li} \quad \textbf{Ganqu Cui} \quad \textbf{Ning Ding} \quad \textbf{Xu Han} \\ \vspace*{2mm}\textbf{Yuan Yao}\thanks{Corresponding authors.} \quad\textbf{Zhiyuan Liu$^*$} \quad\hspace{-2mm}\textbf{Maosong Sun$^*$}\\
\vspace*{2mm}
MiniCPM-V Team, OpenBMB\\
\vspace*{2mm}
\texttt{yiranytianyu@gmail.com \quad yaoyuanthu@gmail.com}\\
\vspace{1mm}
\github \href{https://github.com/openbmb/MiniCPM-V}{{\text{MiniCPM-V 4.5 Code}}} 
\hspace{10mm} 
\huggingface \href{https://huggingface.co/openbmb/MiniCPM-V-4_5}{{\text{MiniCPM-V 4.5 Model}}} 
}

\definecolor{myred}{rgb}{1.0, 0.25, 0}

\usepackage{fancyhdr}

\begin{document}

\maketitle

\begin{abstract}
Multimodal Large Language Models (MLLMs) are undergoing rapid progress and represent the frontier of AI development. However, their training and inference efficiency have emerged as a core bottleneck in making MLLMs more accessible and scalable. To address the challenges, we present MiniCPM-V 4.5, an 8B parameter model designed for high efficiency and strong performance. We introduce three core improvements in model architecture, data strategy and training method: a unified 3D-Resampler model architecture for highly compact encoding over images and videos, a unified learning paradigm for document knowledge and text recognition without heavy data engineering, and a hybrid reinforcement learning strategy for proficiency in both short and long reasoning modes. Comprehensive experimental results in OpenCompass evaluation show that MiniCPM-V 4.5 surpasses widely used proprietary models such as GPT-4o-latest, and significantly larger open-source models such as Qwen2.5-VL 72B. Notably, the strong performance is achieved with remarkable efficiency. For example, on the widely adopted VideoMME benchmark, MiniCPM-V 4.5 achieves state-of-the-art performance among models under 30B size, using just 46.7\% GPU memory cost and 8.7\% inference time of Qwen2.5-VL 7B. 
\end{abstract}

\section{Introduction}

Multimodal Large Language Models (MLLMs)~\citep{vteam2025glm45vglm41vthinkingversatilemultimodal,kimiteam2025kimivltechnicalreport,coreteam2025mimovltechnicalreport,openai2025gpt4olatest,bai2023qwenvlversatilevisionlanguagemodel,yao2024minicpm,bai2025qwen25vltechnicalreport} are advancing rapidly the frontier of artificial intelligence, enabling machines to deeply understand and reason over different modalities such as text and images.
However, as MLLMs evolve, the cost of data engineering, training, and inference also increases heavily.
Addressing this efficiency challenge is now a central focus of both research and industry~\citep{yao2024minicpm,lu2025ovis25technicalreport,zhu2025internvl3,zhang2025flash,gemini2024}, essential for making capable MLLMs more accessible and scalable.

We decompose this efficiency problem into three core aspects: 
(1) \textbf{Model Architecture}. 
A primary efficiency bottleneck in MLLMs comes from the large number of visual tokens for high-resolution image encoding, which brings heavy computation overhead for visual encoders and LLMs. The problem is even exacerbated in video understanding, where existing models can take thousands of tokens to encode a short and low-resolution video, even when sampling at a low frame rate. For example, processing a 6-second, 2-fps video at a resolution of just 448$\times$448 requires 1,536 tokens for Qwen2.5-VL~\citep{bai2025qwen25vltechnicalreport}, and 3,072 tokens for InternVL3~\cite{zhu2025internvl3}.
Such long visual token sequences lead to prohibitive training and inference costs in GPU memory and computation speed. 
(2) \textbf{Training Data}. As we quickly run out of new knowledge from traditional web page data, a new cornerstone of modern MLLMs is harnessing high-quality multimodal knowledge from documents~\citep{vteam2025glm45vglm41vthinkingversatilemultimodal,kimiteam2025kimivltechnicalreport}, such as scientific papers and textbooks. These documents are often stored as PDFs, containing multi-disciplinary knowledge in various domains and organized in diverse layouts of interleaved texts, images, and tables.
However, most methods depend on brittle external parsing tools to convert document files into interleaved image-text sequences for training. 
These tools often fail in complex layouts, leading to either errors in knowledge learning or heavy data engineering efforts to fix the failure cases.
(3) \textbf{Training Methods}. Reinforcement Learning (RL) has shown promise in improving complex reasoning capabilities by enabling a step-by-step explicit thinking process before providing the final answer~\citep{deepseekai2025deepseekr1incentivizingreasoningcapability, vteam2025glm45vglm41vthinkingversatilemultimodal}. However, this performance gain often comes at the expense of extreme verbosity. Even for simple tasks such as identifying obvious objects, most existing thinking models produce excessively long outputs, inducing poor efficiency in both training and inference. For example, on the comprehensive Opencompass benchmark, the hybrid strategy requires only 33.3\% long reasoning samples to match the peak long reasoning performance of training exclusively in single mode.

To address the challenges, MiniCPM-V 4.5 introduces three key improvements in model architecture, data strategy, and training method: (1) \textbf{Unified 3D-Resampler for Compact Image and Video Encoding.} Previous MiniCPM-V series models~\cite{yao2024minicpm} exhibit high compression rates (e.g., 4$\times$ compared with most MLLMs) for high-resolution images via 2D-Resamplers~\cite{bai2023qwenvlversatilevisionlanguagemodel,guo2024llava}. To further address the architectural inefficiency of video processing, we extend the 2D-Resampler to a 3D-Resampler that jointly compresses spatial-temporal information for videos. This module can encode a 6-second, 2-fps, 448$\times$448 resolution video into only 128 visual tokens, achieving a 12$\times$-24$\times$ reduction in token cost compared to representative MLLMs~\cite{bai2025qwen25vltechnicalreport,zhu2025internvl3}, enabling efficient high frame rate and long video understanding, and unified encoding for images as well. 
(2) \textbf{Unified Learning Paradigm for Document Knowledge and OCR.} 
We propose a learning paradigm that enables the model to accurately acquire knowledge directly from document images, eliminating the need for fragile external parsers. By dynamically corrupting text regions in documents with varying noise levels and asking the model to reconstruct the text, the model learns to adaptively and properly switch between accurate text recognition (when text is roughly visible) and multimodal context-based knowledge reasoning (when text is heavily corrupted). 
(3) \textbf{Hybrid Strategy for Post-Training.} Unlike prior models that optimize for a single long reasoning mode~\citep{kimiteam2025kimivltechnicalreport, vteam2025glm45vglm41vthinkingversatilemultimodal}, we develop a hybrid RL post-training strategy to support both short reasoning mode for efficient usage and long reasoning mode for complex tasks.
In RL training, we randomly alternate between the two modes during the rollout process for joint optimization. 
This approach not only enables flexible control over the short and long reasoning modes but also allows for mutual performance enhancement. In experiments, we can achieve better reasoning performance with fewer training samples for both modes.

Comprehensive experimental results in OpenCompass evaluation show that MiniCPM-V 4.5 outperforms widely used proprietary models such as GPT-4o-latest~\citep{openai2025gpt4olatest}, and significantly larger open-source models such as Qwen2.5-VL 72B~\citep{bai2025qwen25vltechnicalreport}. Notably, the strong performance is achieved with remarkable efficiency. For example, powered by the efficient unified 3D-Resampler, MiniCPM-V 4.5 achieves equivalent performance on VideoMME~\citep{fu2025videommefirstevercomprehensiveevaluation} using only 9.9\% of the inference time of prior state-of-the-art MLLMs~\citep{vteam2025glm45vglm41vthinkingversatilemultimodal}. Based on the hybrid post-training strategy, MiniCPM-V 4.5 excels in both short and long reasoning modes, outperforming concurrent thinking models~\citep{coreteam2025mimovltechnicalreport, vteam2025glm45vglm41vthinkingversatilemultimodal} on OpenCompass evaluation while using only 42.9\%-68.2\% inference time. 

In summary, our contributions are as follows:

\begin{itemize}[leftmargin=1em, itemsep=0pt, topsep=-1pt, partopsep=0pt, partopsep=0pt]
    \item We open-source MiniCPM-V 4.5, an efficient and strong MLLM that supports efficient high frame rate and long video understanding, controllable hybrid reasoning, robust OCR, and strong document parsing capabilities.
    \item We introduce three key improvements: a unified 3D-Resampler for efficient image and video encoding, a unified paradigm for document knowledge and OCR learning, and a hybrid strategy for post-training that enhances both performance and efficiency.
    \item Comprehensive experiments demonstrate the effectiveness of the proposed technical improvements and the performance of MiniCPM-V 4.5.
\end{itemize}

\begin{figure}
    \centering
    \includegraphics[width=\linewidth]{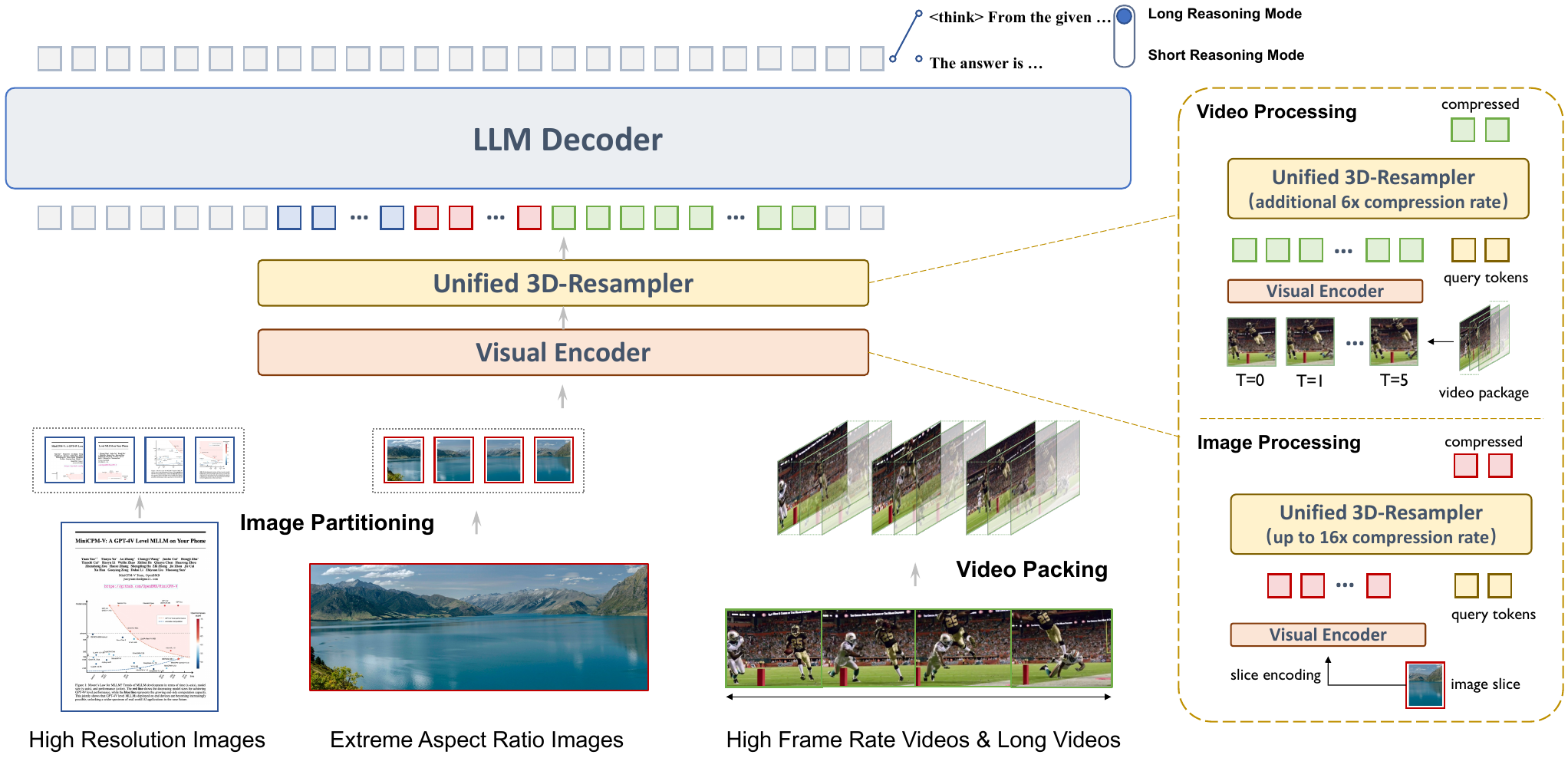}
    \caption{An overview of the MiniCPM-V 4.5 architecture. The model processes diverse visual inputs, such as high-resolution images and high frame rate videos. After the image partitioning and video packing processes, these inputs are encoded by a visual encoder and then fed into the unified 3D-Resampler. This module efficiently compresses both image and video features into a compact token sequence (achieving up to 16$\times$ compression rate for images and an additional 6$\times$ for videos), which is then processed by the LLM decoder. The decoder can generate responses in two distinct styles: a concise, short reasoning mode or a step-by-step, long reasoning mode.}
    \label{fig:framework}
\end{figure}

\section{Approach}

In this section, we describe the methodology of MiniCPM-V 4.5, including the model architecture and the recipes for pre-training, SFT, and RL.

\subsection{Architecture}

As shown in Figure~\ref{fig:framework}, the architecture of MiniCPM-V 4.5 comprises three main modules: (1) A lightweight visual encoder that flexibly handles high-resolution images with a special partitioning strategy. 
(2) A unified 3D-Resampler that encodes images and videos into compact features, exploiting temporal redundancies in visual information. 
(3) An LLM decoder that understands images, videos, and text, and generates text outputs. 

\subsubsection{The Unified 3D-Resampler}

To tackle the image and video encoding efficiency bottleneck in MLLMs, we extend the 2D-Resampler to a 3D-Resampler that jointly compresses spatial-temporal information for videos. In this way, we achieve a 6$\times$ temporal compression rate by leveraging the temporal redundancy of consecutive multiple video frames.

\textbf{Image Processing.} 
To handle high-resolution images in any aspect ratio, we adopt the LLaVA-UHD~\citep{guo2024llava} image partitioning strategy. 
For each image, we estimate the ideal number of slices from the input resolution and choose the partition whose per-slice resolution deviates least from the visual encoder pretraining setting. We then use learnable queries augmented with 2D spatial positional embeddings to produce a fixed-length sequence for each slice through cross-attention. Most existing MLLMs~\cite{bai2025qwen25vltechnicalreport,zhu2025internvl3,vteam2025glm45vglm41vthinkingversatilemultimodal} adopt MLP and pixel unshuffle operation for visual compression, and typically require visual 256 tokens for encoding a 448$\times$448 image. Leveraging the flexibility of resampler architecture, by choosing a small number of query tokens, MiniCPM-V can achieve a significantly higher compression rate for visual tokens (e.g., 64 tokens for a 448$\times$448 image) while maintaining good performance. 

\textbf{Video Processing.} 
To handle the significant redundancy in video data, we employ a joint spatial-temporal compression strategy for higher compression rates. 
For each video, we first split it into packages along the temporal dimension, where each package contains adjacent frames. Intuitively, the video frames within the same package typically share highly redundant visual information, which can be identified and compressed when jointly modeled. To this end, we resample the frame features from the visual encoder in each package into a fixed-length feature sequence through cross-attention. We augment the learnable queries with both 2D spatial positional embedding, as used in image encoding, and temporal positional embedding. The final video representation is obtained by concatenating the token sequences from all packages. We sample at most 1080 frames per video at a maximum frame rate of 10. 
During training, the package size and frame rate are randomly augmented to improve robustness. This design also provides flexibility at inference time, allowing these hyperparameters to be adjusted to meet the demands of diverse scenarios and devices. 

Based on the 3D-Resampler, MiniCPM-V 4.5 can achieve 96$\times$ compression rate for video tokens, where 6,448$\times$448 video frames can be jointly compressed into 64 video tokens (normally 1,536-3,072 tokens for most MLLMs). This means that the model can perceive significantly more video frames without increasing the LLM inference cost, which brings strong high-frame-rate video understanding and long video understanding capabilities.

\textbf{Training Efficiency.} 
Thanks to the flexibility of the resampler mechanism (agnostic to input shape), we can use the same 3D-Resampler for unified visual encoding over images and videos. This means that image and visual encoding share the same architecture and weights, and therefore, we can achieve the extension from 2D-Resampler to 3D-Resampler efficiently via a lightweight SFT stage. Moreover, this also facilitates efficient knowledge transfer from images to videos. For example, we observe reasonable video OCR capability in MiniCPM-V 4.5, although we did not specifically collect such training data.

\begin{tcolorbox}[
    colframe=takeaway,
    colback=white,
    coltitle=takeawayTitle,
]
\textcolor{takeawayTitle}{\textbf{Takeaway}}
\\Joint spatial–temporal compression can enable higher visual compression rates. A unified architecture can be more efficiently adapted with minimal additional training and facilitates knowledge transfer from images to videos.
\end{tcolorbox}

\subsection{Pre-training}
\label{sec:pretrain}
Our pre-training process aims to systematically build the model's foundational capabilities through a progressive, multi-stage strategy. This involves a carefully curated data composition and a novel unified paradigm for document knowledge and OCR learning.

\subsubsection{Pre-training Strategy}

The pre-training comprises three progressive stages. 
Each stage strategically unfreezes different model components and introduces increasingly complex data to optimize learning efficiency. 

\textbf{Stage 1.} We begin with a warm-up stage, training only the 2D-Resampler module while all other components remain frozen. This stage uses image-caption data to establish an initial alignment between visual and language modalities with minimal training cost.

\textbf{Stage 2.} We then unfreeze the vision encoder to enhance the perceptual foundation capability. This stage consumes OCR-rich data and image-caption data. Since the data in this stage may lack the fluency or quality required for language modeling, the LLM decoder remains frozen in this stage. 

\textbf{Stage 3.} With the cross-modal bridge in place and the perceptual foundation set, the final stage trains all model parameters end-to-end using our highest quality data, including text-only corpora, image-text interleaved samples, videos, and a curated subset from earlier stages. At this point, we unfreeze the LLM decoder to fully exploit the knowledge and skills in data, encompassing multi-image reasoning and temporal understanding. We adopt the Warmup-Stable-Decay learning rate scheduler~\citep{hu2024minicpm}. During the decay phase, we gradually add more high-quality instructions and knowledge-intensive data.

\subsubsection{Pre-training Data}

\textbf{Image Caption Data.} 
We combine large-scale public datasets (LAION-2B~\citep{schuhmann2022laion}, COYO~\citep{kakaobrain2022coyo-700m}, etc.) with curated Chinese image-text pairs crawled from the web.
We filter out low-resolution images and remove irrelevant image-text pairs with CLIP~\cite{Radford2021LearningTV}. 
To enrich alt-text descriptions, we employ a Capsfusion-based~\citep{yu2024capsfusion} re-captioning process on a subset to generate fluent and factually complete captions. 
In this way, we formulate the valuable world knowledge in raw captions into more fluent natural language. 
We employ an MLLM to tag images with concept labels and ensure a balanced distribution across languages and long-tail concepts.

\textbf{Image-Text Interleaved Data.}
Sourced from Common Crawl, OmniCorpus~\citep{li2024omnicorpus}, and MINT-1T~\citep{awadalla2024mint}, image-text interleaved data is crucial for in-context learning and multi-image understanding capabilities.
We apply filtering to ensure quality, removing samples with broken images or imbalanced image-text ratios. 
We further use relevance filtering to ensure meaningful multimodal associations, and employ knowledge density filtering to select a high-quality subset for the final decay phase of pre-training.

\textbf{OCR Data.}
We synthesize OCR data to enhance the basic text recognition capability during the early pre-training stage. We render text on natural scenes with various combinations of color and font following~\cite{gupta2016synthetic}, and also render real-world HTML sources into images.

\textbf{Document Data.}
We collect documents, including scientific papers, academic reports, textbooks, etc., from the web. This data exhibits high knowledge density and contains visually complex layouts.

\textbf{Video Caption Data.} 
We aggregate several public datasets (WebVid~\cite{Bain21}, Vript~\cite{yang2024vript}, OpenVid~\cite{nan2024openvid}) and supplement them with detailed video captions.  
This diverse collection supports the development of temporal visual reasoning capabilities essential for video comprehension.

\subsubsection{Unified Paradigm for Document Knowledge and OCR Learning}

\begin{figure}[t]
\vspace{-4mm}
    \centering
    \includegraphics[width=\linewidth]{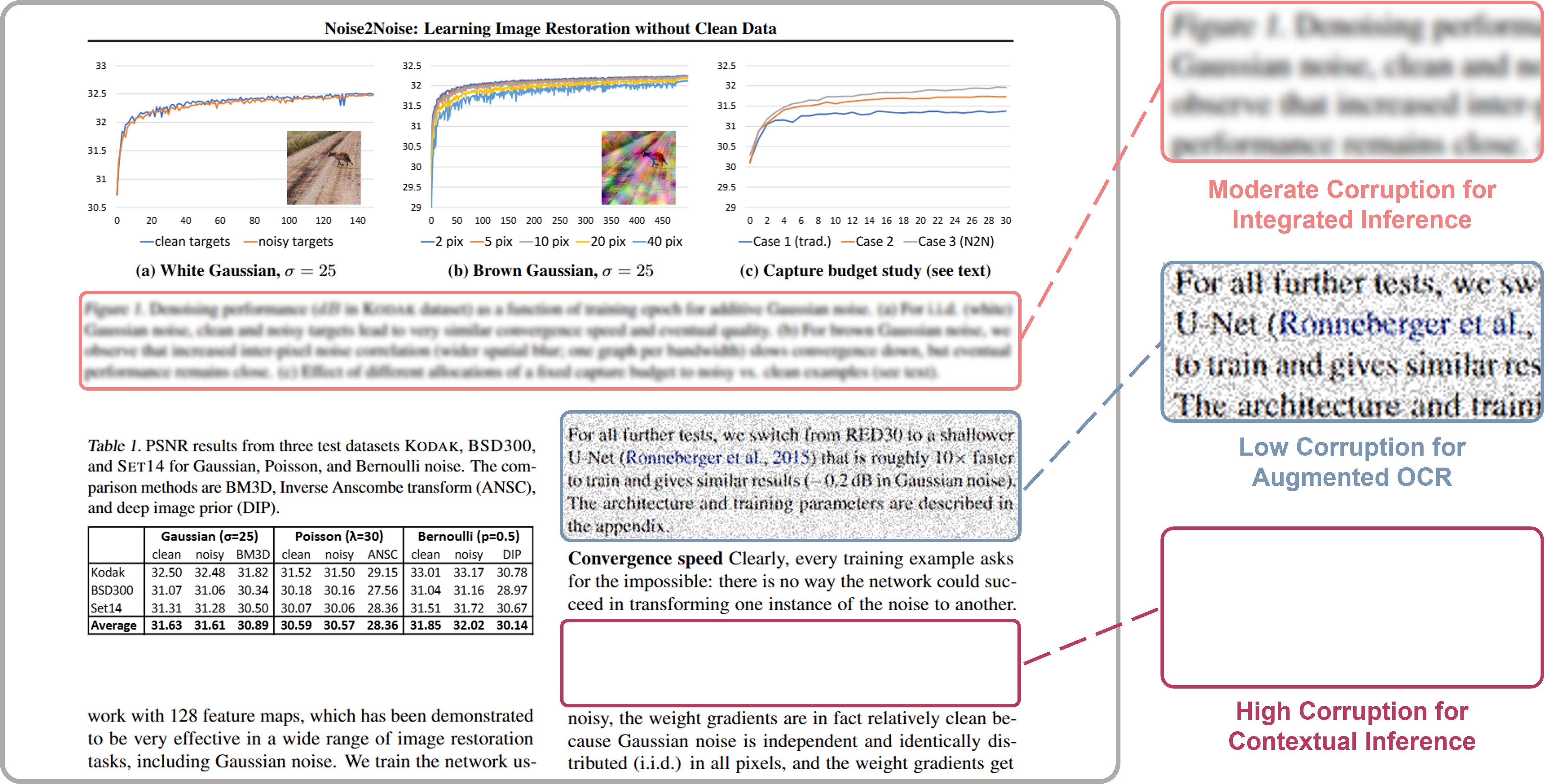}
    \caption{Unified paradigm for document knowledge and OCR learning via dynamic visual corruption. We create a spectrum of training tasks through varied corruption levels: low corruption preserves readability to learn robust OCR, high corruption forces the model to perform contextual inference, and moderate corruption requires integrated inference from visual clues and context.}
    \label{fig:ocr}
\end{figure}

Documents, such as scientific papers, textbooks, and web pages, are vital resources for learning diverse layouts and acquiring multi-disciplinary knowledge in various domains. 
However, most MLLMs depend on brittle external parsers to convert document PDFs into an interleaved image-text sequence for training. Such a noisy and inefficient process often introduces structural errors or requires heavy data engineering efforts to fix the failure cases. 

Another challenge for OCR learning is that, while stronger image augmentation can create more diverse and harder samples, leading to more robust OCR capabilities, over-augmentation can make the texts indistinguishable. Forcing the model to produce the ground truth text from such indistinguishable visual input typically leads to hallucination problems. Therefore, previously, we could only afford a small and safe augmentation level.

To overcome both challenges, we propose a unified training paradigm that learns directly from document images, using their original text as ground truth. 
Our key insight is that the key difference between document knowledge acquisition and text recognition is the visibility of the text in images.
We unify both capabilities into a single learning objective: predicting original text from corrupted document images. 
By dynamically corrupting text regions with varying corruption levels, the model learns to adaptively and properly switch between precise text recognition (when text is distinguishable) and multimodal context-based knowledge reasoning (when text is heavily obscured or masked), as illustrated in Figure~\ref{fig:ocr}. This eliminates reliance on fragile parsers and prevents hallucinations from over-augmented OCR data.

Specifically, for each document, we treat a subset of its text regions as training ground truth. We then stochastically apply different levels of corruption to each region, essentially creating different training tasks:

\begin{enumerate}[leftmargin=*]
\item \textbf{Low Corruption (Augmented OCR).} When mild noise is applied to a text region, the texts are still recognizable, and the model could effectively predict them via text recognition.
\item \textbf{Moderate Corruption (Integrated Inference).} When heavy noise is applied to the text region, individual characters become highly ambiguous and unreliable for recognition. The model must therefore learn to integrate the noisy visual cues from the corrupted region with the high-level document context and its internal knowledge to reconstruct the original text.
\item \textbf{High Corruption (Contextual Inference and Document Knowledge Learning).} With the text region completely masked out, the model cannot rely on character-level cues to predict the missing content. Consequently, the model is forced to infer the information only from the multimodal context and its internal knowledge, including other text, layout structures, charts, tables, and images. This directly cultivates document-level understanding.
\end{enumerate}

This unified approach yields a more efficient and resilient learning process. By learning directly from the document's visual and textual structure, we avoid building complex document parsing pipelines and prevent potential noise introduced by fragile parsers.
Furthermore, this paradigm allows us to fluidly combine knowledge learning and OCR objectives within the same training batch, maximizing data utility and producing a single, versatile model adept at a wide range of document understanding tasks.

\begin{tcolorbox}[
    colframe=takeaway,
    colback=white,
    coltitle=takeawayTitle,
]
\textcolor{takeawayTitle}{\textbf{Takeaway}}

\begin{enumerate}[leftmargin=*]
\item Foundation skills can be built on imperfect heterogeneous data sources by selectively freezing parameters. 
\item Simple dynamic visual corruption on document image text can effectively unify knowledge learning, robust OCR and contextual inference into a single learning objective.
\end{enumerate}

\end{tcolorbox}

\subsection{Supervised Fine-tuning}

The Supervised Fine-Tuning (SFT) stage aims to activate the model's capability on a broad range of tasks and prepares for reinforcement learning. 
Moreover, we extend the 2D-Resampler to a unified 3D-Resampler at this stage to enhance the compression efficiency of video data.

\subsubsection{Supervised Fine-tuning Strategy}
We first train the general interaction abilities, and then cultivate specialized skills for advanced reasoning and temporal understanding. 

\textbf{Stage 1: General SFT.} 
This stage aims to activate the broad knowledge acquired during pre-training and align it with human instructions. 
By fine-tuning on a diverse mixture of high-quality instruction-response data, the model develops proficiency in multimodal interaction. 
To prevent degradation of text-only performance and improve training stability, we include 10\% high-quality text-only data in the training mixture. 

\textbf{Stage 2: Long-CoT \& 3D-Resampler.}
Building on versatile foundations from the previous stage, we then cultivate specialized skills to support long reasoning mode, high frame rate, and long video understanding. 
First, we unlock advanced reasoning by introducing the Long-CoT warm-up instructions into the SFT data. This encourages the model to perform an explicit step-by-step thinking process, incorporating cognitive patterns such as reflection and backtracking, which are vital for the long reasoning mode.
Second, we enhance its temporal understanding by upgrading the architecture from 2D to 3D-Resampler and introducing high frame rate and long video data. Due to the unified design, we find that such an upgrade can be achieved efficiently with a small amount of high-quality video data.

\subsubsection{Supervised Fine-tuning Data}

\textbf{STEM Data.} 
To enhance STEM reasoning, we curate a dataset of high-school and higher multidisciplinary problems from online educational websites, covering physics, chemistry, biology, finance, computer science, etc. 
To ensure the data quality, we implement a two-stage filtering process. First, we only keep samples that exhibit high visual dependency (i.e., not solvable without image information). Second, we perform a consistency check to validate the correctness of the answers. 
For each remaining sample, we perform rejection sampling with a powerful MLLM to collect a clean reasoning process.

\textbf{Long-tail Knowledge Data.} To address the long-tail problem where models often fail on less common topics, we incorporate long-tail knowledge from Wikipedia ~\citep{wikipedia_site_en} to synthesize high-quality multimodal instruction-following data.
Specifically, for each entity page, we construct multimodal instructions and answers using strong MLLMs and keep samples with high visual dependency.

\textbf{Long-CoT Data.} 
Long-CoT data enables the model to acquire the necessary reasoning patterns for the long reasoning mode.
Our data comes from OpenThoughts~\citep{guha2025openthoughts} and an in-house pipeline.
We identify challenging prompts by filtering for those on which our early-stage models struggle. 
Our pilot studies show that focusing on challenging problems is the key to developing robust reasoning capabilities rather than memorizing trivial patterns.
Each response then undergoes a multistage validation: we verify its correctness, assess trustworthiness with claim-level factual verification using RLAIF-V~\citep{yu2024rlaifvopensourceaifeedback}, and filter out meaningless repetition. 
Finally, validated responses are augmented through rewriting to enhance diversity.

\begin{tcolorbox}[
    colframe=takeaway,
    colback=white,
    coltitle=takeawayTitle,    
]
\textcolor{takeawayTitle}{\textbf{Takeaway}}
\\Filtering out easy prompts and focusing on challenging problems is crucial for effective Long-CoT warm-up. 
\end{tcolorbox}

\subsection{Reinforcement Learning}

The RL stage aims to enhance reasoning performance, enable controllable reasoning modes, and improve trustworthiness. To provide efficient general-domain rewards, we combine rule-verified rewards for straightforward cases with general probability-based rewards from RLPR~\citep{yu2025rlprextrapolatingrlvrgeneral} for complex answers and add a calibrated preference reward. A hybrid RL strategy is adopted to allow flexible switch between short and long reasoning modes. We further integrate RLAIF-V~\citep{yu2024rlaifvopensourceaifeedback} to reduce hallucinations.

\subsubsection{Reinforcement Learning Data}

Our RL data contains high-quality samples that span four key domains. Each subset underwent a rigorous, human-in-the-loop cleaning and deduplication process.

\textbf{Mathematics.} We collect multimodal math problems from academic sources~\citep{gao2025gllavasolvinggeometricproblem,deng2024r,lindström2022clevrmathdatasetcompositionallanguage}, which require the integration of visual perception and logical reasoning. 
We observe that many open-source datasets contain severe label errors and adopt a thorough cleaning process to produce the final high-quality set. 

\textbf{Documents, Tables, and Charts.} To improve reasoning on perceptually complex scenarios, we curate a diverse mix of real-world datasets~\citep{gupta-etal-2020-infotabs,lu2023dynamic,pasupat-liang-2015-compositional,Chen2020TabFact,zhu-etal-2021-tat} and synthetic datasets~\citep{ebrahimi2018figureqa,li-etal-2024-multimodal-arxiv,Kafle_2018_CVPR} to improve the coverage of domains.

\textbf{General Reasoning.} To further improve general reasoning capabilities, we assemble a diverse collection of problems covering logical and multi-disciplinary reasoning tasks from VisualWebInstruct~\citep{visualwebinstruct} and additional web resources. These data exhibit a more complex reference answer style; many of the problems have more than one sub-question.

\textbf{Instruct Following.} We incorporate text-only instructions from the Llama-Nemotron-Post-Training Dataset \citep{bercovich2025llamanemotronefficientreasoningmodels} and the MulDimIF dataset \citep{MulDimIF}. We observe that the instruction-following improvement generalizes well to multimodal instructions.

\subsubsection{Reward Quality Control}

The efficacy of RL is highly dependent on data quality. Thus, we implement meticulous quality control processing, focusing on two distinct aspects: 

\textbf{Label Accuracy.} Incorrect labels can introduce flawed supervision signals. For each dataset, we maintain a small subset to inspect the label accuracy and conduct a human-in-the-loop cleaning process to keep a high label accuracy.

\textbf{Rewarding Accuracy.} Verifying model-generated responses in the general domain is a nontrivial challenge. Hand-crafted rules struggle to tackle the complexity of natural language. 
To address this, we dynamically apply the most suitable validation method for each case.
For straightforward answers containing only a few tokens, we employ a rule-based verification system, achieving 98\% reward accuracy. 
For complex natural language answers where rules are brittle (e.g., those containing specific units or longer phrasing), we use the more robust probability-based rewards of RLPR~\citep{yu2025rlprextrapolatingrlvrgeneral}.

\textbf{Rewarding Coverage.} To complement these accuracy-focused signals, we integrate a reward model to provide a dense preference-aligned signal that guides the model towards higher-quality human-like responses. We apply the reward model to only the final answer part for the long reasoning mode to avoid the out-of-distribution problem.

\subsubsection{Hybrid Reinforcement Learning}

We adopt a controllable hybrid reasoning design for our RL model: a short reasoning mode for quick answers and a long reasoning mode that emits explicit step-by-step traces for complex problems. 
Mode switching is controlled by prompts. 
Both behaviors are initialized during SFT and then optimized jointly via hybrid RL, where rollouts randomly alternate between the two modes. 

We apply GRPO~\citep{shao2024deepseekmathpushinglimitsmathematical} to optimize the model with these rollouts and remove the KL and entropy loss to improve stability.
This training schedule not only preserves the efficiency of short responses while retaining complex reasoning capabilities, but also fosters cross-generalization, where reasoning capabilities learned in one mode can transfer to improve the other mode.

\subsubsection{Reward Shaping}

We design the reward shaping strategy to balance task capability, human preference, and training stability. 
The final reward signal is a weighted composite of four components: an accuracy reward $R_\textbf{acc}$, a format reward $R_\textbf{format}$, a repetition penalty reward $R_\textbf{rep}$, and a preference reward $R_\textbf{rm}$. 
The preference reward is derived from an auxiliary RM trained with human preference data~\citep{wang2025skyworkvlrewardeffectivereward}. However, directly applying RMs in the long reasoning mode yields unsatisfactory results since standard RMs struggle to evaluate the out-of-distribution long reasoning chains, leading to worse alignment and training instability, which is also confirmed in our preliminary experiments.

To address this, we adopt a selective application strategy. The RM scores only the final answer part of the response, completely bypassing the explicit thinking steps. This provides a stable, dense reward signal that aligns with human preferences without incorrectly penalizing complex reasoning paths.
The final reward is calculated as follows.

\begin{equation}
    R = R_\textbf{acc} + R_\textbf{format} + R_\textbf{rep} + \frac{1}{2}\tilde{R}_\textbf{rm}.
\end{equation}
Here, $\tilde{R}_\textbf{rm}$ is the standardized preference reward score computed using $\frac{R_\textbf{rm} - \bar{R}_\textbf{rm}}{\sigma(R_\textbf{rm})}$, where $\bar{R}_\textbf{rm}$ and $\sigma(R_\textbf{rm})$ represent the average and standard deviation of raw reward scores from responses sampled with the same prompt.

\subsubsection{RLAIF-V}

Visual hallucinations remain a critical limitation for MLLMs, particularly in applications requiring high reliability. 
To address this challenge, we integrate RLAIF-V~\citep{yu2024rlaifvopensourceaifeedback} to make the responses more factually grounded to the visual input through alignment from scalable AI feedback.
Notably, we extend this approach to video inputs, where hallucination problems are especially pronounced.

\textbf{Response Sampling.} We first sample multiple responses from the policy model under the same generation condition. 
This strategy ensures focused evaluation of factual accuracy, avoiding distributional mismatches between models.

\textbf{Feedback Collection.} We begin by decomposing complex responses into verifiable atomic claims, where each claim is independently validated. This transforms the complex long response evaluation into simpler claim-level verification, addressing the inherent challenge of holistic assessment and improving the precision of factual evaluation. Preference pairs are then constructed based on aggregated claim verification scores, where responses containing fewer factual errors are preferred.

\textbf{Preference Learning.} 
The resulting preference dataset,  encompassing both image and video modalities, is used to train the model with DPO~\citep{rafailov2023direct}. This stage proves particularly effective for visual tasks where factual accuracy is paramount, without compromising response quality or natural language fluency.

\begin{tcolorbox}[
colframe=takeaway,
colback=white,
coltitle=takeawayTitle,
]
\textcolor{takeawayTitle}{\textbf{Takeaway}}
\begin{enumerate}[leftmargin=*]
\item Combining rule-based reward for simple responses and probability-based reward for complex natural language responses enables a reliable reward system for diverse tasks. 
\item Hybrid RL enables cross-mode generalization between long and short reasoning modes. 
\end{enumerate}
\end{tcolorbox}

\section{Experiments}
In this section, we empirically evaluate the performance of MiniCPM-V 4.5, and the effectiveness of the proposed methods. 

\subsection{Baselines and Benchmarks}

We compare with various strong baseline models: (1) state-of-the-art open-source models, represented by Qwen2.5-VL 72B~\citep{bai2025qwen25vltechnicalreport}; (2) strong models of comparable parameter sizes, encompassing parameter-matched competitors such as InternVL3~\citep{zhu2025internvl3} (8B), and GLM-4.1V~\citep{vteam2025glm45vglm41vthinkingversatilemultimodal} (9B); and (3) frontier proprietary models such as the latest GPT-4o~\citep{openai2025gpt4olatest}.

Our evaluation encompasses several key areas of multimodal capabilities:

\textbf{STEM} includes mathematics and science-oriented benchmarks such as MMMU~\citep{yue2024mmmu}, MathVista~\citep{lu2024mathvista}, AI2D~\citep{kembhavi2016diagram}, MathVerse~\citep{zhang2024mathverse}, LogicVista~\citep{xiao2024logicvista}, and EMMA~\citep{hao2025can}, designed to evaluate logical reasoning, mathematical problem-solving, and scientific understanding capabilities.

\textbf{Document, OCR \& Chart} covers OCR-related tasks through OCRBench~\citep{liu2024ocrbench}, ChartQA~\citep{masry2022chartqa}, TextVQA~\citep{singh2019textvqa}, DocVQA~\citep{mathew2021docvqa}, and OmniDocBench~\citep{ouyang2025omnidocbenchbenchmarkingdiversepdf}, testing ability to extract, interpret, and reason about textual information in various visual contexts, including documents and charts.

\textbf{Hallucination} evaluates model reliability through HallusionBench~\citep{guan2024hallusionbench}, ObjHalBench~\citep{rohrbach2018object}, and MMHal-Bench~\citep{sun2023aligning}, measuring the tendency to generate false or inconsistent information.

\textbf{Multi-Image \& Real-World \& Instruction Following} includes Mantis~\citep{jiang2024mantis}, MMT-Bench~\citep{ying2024mmt}, RealWorldQA~\citep{xai2024grok15v}, and MM-IFEval~\citep{ding2025mm}, assessing performance on complex scenarios involving multiple images, real-world understanding, and instruction following.

\textbf{Video Understanding} encompasses Video-MME~\citep{fu2024videomme}, LVBench~\citep{wang2024lvbench}, MLVU~\citep{zhou2025mlvu}, LongVideoBench~\citep{wu2024longvideobench}, MotionBench~\citep{hong2024motionbench}, and FavorBench~\citep{tu2025favorbenchcomprehensivebenchmarkfinegrained}, evaluating temporal reasoning and dynamic visual comprehension across various video tasks.

\textbf{Comprehensive Multimodal Understanding} includes benchmarks such as OpenCompass~\citep{2023opencompass}, MMVet~\citep{yu2023mm}, MMStar~\citep{chen2024we}, MME~\citep{fu2024mmecomprehensiveevaluationbenchmark}, and MMBench V1.1~\citep{liu2024mmbench}, which assess general vision-language comprehension across diverse task types. 
Within the OpenCompass average, we use the long reasoning mode for 5 benchmarks, including MMStar, MMVet, HallusionBench, MathVista, and MMMU.

\begin{table*}[t]
\centering
\tiny
\begin{threeparttable}
\definecolor{headlineblue}{HTML}{F4FAFF}
\definecolor{cellblue}{HTML}{F4FAFF}
\colorlet{headlinecolor}{headlineblue!100} 
\colorlet{highlightcolor}{cellblue!100} 
\renewcommand{\arraystretch}{3.0}
\resizebox{\textwidth}{!}{%
\begin{tabular}{>{\fontsize{10pt}{12pt}\selectfont}l >{\fontsize{10pt}{12pt}\selectfont}l |>{\fontsize{10pt}{12pt}\selectfont}c |*{4}{>{\fontsize{10pt}{12pt}\selectfont}c} *{4}{>{\fontsize{10pt}{12pt}\selectfont}c}}

\toprule[2pt]
\tablehead{\scalebox{1.25}{Task}} & \tablehead{\scalebox{1.25}{Benchmark}} & 
\cellcolor{headlinecolor}\tablehead{\scalebox{1.25}{MiniCPM-V 4.5}} & \tablehead{\scalebox{1.25}{Qwen2.5-VL}} & \tablehead{\scalebox{1.25}{Qwen2.5-VL}} & \tablehead{\scalebox{1.25}{InternVL3}} & \tablehead{\scalebox{1.25}{GLM-4.1V}} & \tablehead{\scalebox{1.25}{GPT-4o}} \\

\midrule[2pt]

\scalebox{1.25}{Size} & & \cellcolor{highlightcolor}\scalebox{1.25}{8B} & \scalebox{1.25}{7B} & \scalebox{1.25}{72B} & \scalebox{1.25}{8B} & \scalebox{1.25}{9B} & \scalebox{1.25}{-} \\ 
\scalebox{1.25}{Mode} & & \cellcolor{highlightcolor}\scalebox{1.25}{hybrid} & \scalebox{1.25}{non-thinking} & \scalebox{1.25}{non-thinking} & \scalebox{1.25}{non-thinking} & \scalebox{1.25}{thinking} & \scalebox{1.25}{non-thinking} \\

\midrule[1pt]
\multirow{5}{*}{\scalebox{1.25}{{\raggedright\makecell[l]{Comprehensive \\ Multimodal}}}}
& OpenCompass& \cellcolor{highlightcolor}\scalebox{1.25}{\textbf{77.0$^\dagger$}} & \scalebox{1.25}{70.5} &\scalebox{1.25}{76.1} & \scalebox{1.25}{73.6} & \scalebox{1.25}{76.6} & \scalebox{1.25}{75.4$^\ddagger$}\\
& MMVet & \cellcolor{highlightcolor}\scalebox{1.25}{75.5$^\dagger$} &  \scalebox{1.25}{67.1} & \scalebox{1.25}{76.9} & \scalebox{1.25}{\textbf{81.3}} & \scalebox{1.25}{\hspace{0.5mm}70.5$^\dagger$} & \scalebox{1.25}{76.9$^\ddagger$}\\
& MMStar & \cellcolor{highlightcolor}\scalebox{1.25}{72.1$^\dagger$} &  \scalebox{1.25}{63.9} & \scalebox{1.25}{70.5}& \scalebox{1.25}{68.2}& \scalebox{1.25}{\textbf{72.9}} & \scalebox{1.25}{70.2$^\ddagger$}\\
& MME & \cellcolor{highlightcolor}\scalebox{1.25}{\textbf{2500}} &  \scalebox{1.25}{2347} & \scalebox{1.25}{2483} & \scalebox{1.25}{2415} & \scalebox{1.25}{\hspace{1mm}2466$^\dagger$} & \scalebox{1.25}{2318$^*$}\\
& MMBench V1.1 & \cellcolor{highlightcolor}\scalebox{1.25}{84.2$^\dagger$} &  \scalebox{1.25}{82.6} & \scalebox{1.25}{\textbf{87.8}}& \scalebox{1.25}{81.7}& \scalebox{1.25}{85.3} & \scalebox{1.25}{86.0$^\ddagger$}\\

\midrule[1pt]
\multirow{6}{*}{\scalebox{1.25}{STEM}} & MMMU & \cellcolor{highlightcolor}\scalebox{1.25}{\hspace{1.5mm}67.7$^\dagger$} &  \scalebox{1.25}{58.6} & \scalebox{1.25}{68.2} & \scalebox{1.25}{62.7} & \scalebox{1.25}{68.0} & \scalebox{1.25}{\textbf{72.9$^\ddagger$}} \\
& MathVista & \cellcolor{highlightcolor}\scalebox{1.25}{\hspace{1.5mm}79.9$^\dagger$} &  \scalebox{1.25}{68.2} & \scalebox{1.25}{74.2}& \scalebox{1.25}{71.6}& \scalebox{1.25}{\textbf{80.7}} & \scalebox{1.25}{71.6$^\ddagger$}\\
& AI2D & \cellcolor{highlightcolor}\scalebox{1.25}{\hspace{0.5mm}86.5} &  \scalebox{1.25}{83.9} & \scalebox{1.25}{\textbf{88.5}} & \scalebox{1.25}{85.2} & \scalebox{1.25}{87.9} & \scalebox{1.25}{86.3$^\ddagger$}\\
& MathVerse MINI & \cellcolor{highlightcolor}\scalebox{1.25}{\hspace{2mm}58.8$^\dagger$} &  \scalebox{1.25}{49.2} & \scalebox{1.25}{47.3}& \scalebox{1.25}{39.8}& \scalebox{1.25}{\textbf{68.4}} & \scalebox{1.25}{40.6}\\
& LogicVista & \cellcolor{highlightcolor}\scalebox{1.25}{\hspace{2mm}57.0$^\dagger$} &  \scalebox{1.25}{44.1} & \scalebox{1.25}{55.7} & \scalebox{1.25}{44.1} & \scalebox{1.25}{\textbf{60.4}} & \scalebox{1.25}{52.8}\\
& EMMA & \cellcolor{highlightcolor}\scalebox{1.25}{\hspace{2mm}34.8$^\dagger$} &  \scalebox{1.25}{\hspace{1mm}28.6$^*$} & \scalebox{1.25}{-} & \scalebox{1.25}{-} & \scalebox{1.25}{\textbf{\hspace{1.0mm}35.7$^\dagger$}} & \scalebox{1.25}{32.4}\\

\midrule[1pt]
\multirow{6}{*}{\scalebox{1.25}{\raggedright\makecell[l]{Document, \\ OCR \& Chart}}}
& OCRBench & \cellcolor{highlightcolor}\scalebox{1.25}{\textbf{\hspace{0.5mm}89.0}} &  \scalebox{1.25}{86.4} & \scalebox{1.25}{88.2} & \scalebox{1.25}{88.0} & \scalebox{1.25}{84.2} & \scalebox{1.25}{82.2$^\ddagger$}\\
& ChartQA & \cellcolor{highlightcolor}\scalebox{1.25}{\hspace{0.5mm}87.4} &  \scalebox{1.25}{87.3} & \scalebox{1.25}{\textbf{89.5}} & \scalebox{1.25}{86.6} & \scalebox{1.25}{87.1$^\dagger$}  & \scalebox{1.25}{86.7}\\
& TextVQA & \cellcolor{highlightcolor}\scalebox{1.25}{\hspace{0.5mm}82.2} &  \scalebox{1.25}{84.9} & \scalebox{1.25}{83.5} & \scalebox{1.25}{80.2} & \scalebox{1.25}{79.9$^\dagger$} & \scalebox{1.25}{\hspace{1mm}\textbf{85.6$^*$}}\\
& DocVQA & \cellcolor{highlightcolor}\scalebox{1.25}{\hspace{2mm}94.7$^\dagger$} &  \scalebox{1.25}{95.7} & \scalebox{1.25}{\textbf{96.4}} & \scalebox{1.25}{92.7} & \scalebox{1.25}{93.4$^\dagger$} & \scalebox{1.25}{93.0} \\
& OmniDocBench  (EN) $\downarrow$ & \cellcolor{highlightcolor}\scalebox{1.25}{\textbf{\hspace{2mm}0.175}} &  \scalebox{1.25}{0.316} & \scalebox{1.25}{0.214} & \scalebox{1.25}{0.335$^*$} & \scalebox{1.25}{0.460$^*$} & \scalebox{1.25}{0.233}\\
& OmniDocBench  (ZH) $\downarrow$ & \cellcolor{highlightcolor}\scalebox{1.25}{\textbf{\hspace{2mm}0.253}} &  \scalebox{1.25}{0.399} & \scalebox{1.25}{0.261} & \scalebox{1.25}{0.390$^*$} & \scalebox{1.25}{0.573$^*$} & \scalebox{1.25}{0.399} \\

\midrule[1pt]
\multirow{5}{*}{\scalebox{1.25}{Hallucination}} & HallusionBench & \cellcolor{highlightcolor}\scalebox{1.25}{\hspace{2mm}61.2$^\dagger$} &  \scalebox{1.25}{\hspace{-0.5mm}52.9} & \scalebox{1.25}{54.6}& \scalebox{1.25}{49.9}& \scalebox{1.25}{\textbf{63.2}} & \scalebox{1.25}{57.0$^\ddagger$}\\
& ObjHalBench (CHAIRs) $\downarrow$ & \cellcolor{highlightcolor}\scalebox{1.25}{\textbf{\hspace{2mm}9.3$^\dagger$}} & \scalebox{1.25}{13.7$^*$} & \scalebox{1.25}{\hspace{0.5mm}17.0$^*$} & \scalebox{1.25}{11.3$^*$} & \scalebox{1.25}{12.3$^*$} &\scalebox{1.25}{-}\\
& ObjHalBench (CHAIRi) $\downarrow$ & \cellcolor{highlightcolor}\scalebox{1.25}{\textbf{\hspace{2mm}5.2$^\dagger$}} & \scalebox{1.25}{\hspace{0.5mm}7.7$^*$} & \scalebox{1.25}{8.9$^*$} & \scalebox{1.25}{6.5$^*$} & \scalebox{1.25}{6.4$^*$} &\scalebox{1.25}{-}\\
& MMHal-Bench (Score) & \cellcolor{highlightcolor}\scalebox{1.25}{\textbf{\hspace{2mm}5.0$^\dagger$}}  & \scalebox{1.25}{\hspace{0.5mm}4.1$^*$} & \scalebox{1.25}{4.2$^*$} & \scalebox{1.25}{4.2$^*$} & \scalebox{1.25}{4.6$^*$} &\scalebox{1.25}{-} \\
& MMHal-Bench (Rate)$\downarrow$ & \cellcolor{highlightcolor}\scalebox{1.25}{\textbf{\hspace{2mm}19.4$^\dagger$}}  & \scalebox{1.25}{31.6$^*$} & \scalebox{1.25}{38.2$^*$} & \scalebox{1.25}{24.3$^*$} & \scalebox{1.25}{22.9$^*$} &\scalebox{1.25}{-} \\

\midrule[1pt]
\multirow{4}{*}{\scalebox{1.25}{\raggedright\makecell[l]{Multi-Image \& \\ Real World \& \\ Instruction Following}}}
& Mantis & \cellcolor{highlightcolor}\scalebox{1.25}{\textbf{\hspace{2mm}82.5$^\dagger$}} &  \scalebox{1.25}{\hspace{1mm}74.7$^{*}$} & \scalebox{1.25}{\hspace{0.5mm}81.1$^{*}$} & \scalebox{1.25}{70.1} & \scalebox{1.25}{78.8$^\dagger$} &\scalebox{1.25}{-}\\
& MMT-Bench & \cellcolor{highlightcolor}\scalebox{1.25}{\hspace{1mm}\textbf{68.3}}  & \scalebox{1.25}{63.6} & \scalebox{1.25}{-} & \scalebox{1.25}{65.0} & \scalebox{1.25}{67.6} & \scalebox{1.25}{66.7$^*$} \\
& RealWorldQA & \cellcolor{highlightcolor}\scalebox{1.25}{\hspace{2mm}72.1$^\dagger$} &  \scalebox{1.25}{68.5} & \scalebox{1.25}{75.7} & \scalebox{1.25}{70.8} & \scalebox{1.25}{70.7$^\dagger$} & \scalebox{1.25}{\textbf{76.8$^{*}$}}\\
& MM-IFEval & \cellcolor{highlightcolor}\scalebox{1.25}{\hspace{1mm}66.0} &  \scalebox{1.25}{\hspace{1mm}51.3$^{*}$} & \scalebox{1.25}{\textbf{\hspace{1.0mm}73.8$^{*}$}}& \scalebox{1.25}{\hspace{1mm}53.2$^{*}$}& \scalebox{1.25}{58.4$^\dagger$} & \scalebox{1.25}{\hspace{-1mm}64.6}\\

\midrule[1pt]
\multirow{7}{*}{\scalebox{1.25}{Video Understanding}} & Video-MME (w/o subs) & \cellcolor{highlightcolor}\scalebox{1.25}{\hspace{1.5mm}67.9} & \scalebox{1.25}{65.1} & \scalebox{1.25}{\textbf{73.3}} & \scalebox{1.25}{66.3} & \scalebox{1.25}{68.2} & \scalebox{1.25}{71.9}\\
& Video-MME (w/ subs) & \cellcolor{highlightcolor}\scalebox{1.25}{\hspace{1.5mm}73.5} & \scalebox{1.25}{71.6} & \scalebox{1.25}{\textbf{79.1}} & \scalebox{1.25}{68.9} & \scalebox{1.25}{73.6} &\scalebox{1.25}{77.2}\\
& LVBench & \cellcolor{highlightcolor}\scalebox{1.25}{\textbf{\hspace{1.5mm}50.4}} &  \scalebox{1.25}{45.3} & \scalebox{1.25}{47.3}& \scalebox{1.25}{\hspace{1mm}44.1$^{*}$}& \scalebox{1.25}{44.0} &\scalebox{1.25}{48.9}\\
& MLVU (M-Avg) & \cellcolor{highlightcolor}\scalebox{1.25}{\textbf{\hspace{1.5mm}75.1}} & \scalebox{1.25}{70.2} & \scalebox{1.25}{74.6} & \scalebox{1.25}{71.4} & \scalebox{1.25}{\hspace{0.5mm}72.5$^\dagger$} &\scalebox{1.25}{-}\\
& LongVideoBench (val) & \cellcolor{highlightcolor}\scalebox{1.25}{\hspace{1.5mm}63.9} &  \scalebox{1.25}{56.0} & \scalebox{1.25}{60.7}& \scalebox{1.25}{58.8}& \scalebox{1.25}{\textbf{65.7}} &\scalebox{1.25}{-}\\
& MotionBench & \cellcolor{highlightcolor}\scalebox{1.25}{\textbf{\hspace{1.5mm}59.7}} & \scalebox{1.25}{53.0} & \scalebox{1.25}{58.3} & \scalebox{1.25}{58.1} & \scalebox{1.25}{59.0} &\scalebox{1.25}{58.0}\\
& FavorBench & \cellcolor{highlightcolor}\scalebox{1.25}{\textbf{\hspace{1.5mm}56.0}} &\scalebox{1.25}{42.3} & \scalebox{1.25}{48.1} & \scalebox{1.25}{45.3} & \scalebox{1.25}{\hspace{1mm}51.2$^\dagger$} &\scalebox{1.25}{-}\\
\bottomrule[2pt]
\end{tabular}
}
\caption{
    Evaluation results across diverse vision-language benchmarks. 
    The best performance is marked in \textbf{bold}. $^{*}$~We evaluate officially released checkpoints by ourselves.
$^\dagger$~\hspace{0.2mm}Reasoning mode used, where the average score of three runs is reported for robust evaluation.
$^\ddagger$~\hspace{0.2mm}{GPT-4o-latest evaluation results from OpenCompass. Otherwise GPT-4o-1120 is used in evaluation, since GPT-4o-latest is only accessible via Web API.}
}
\label{tab:general_vqa_comparison}
\vspace{2mm}

\end{threeparttable}
\end{table*}

\subsection{Main Results}

As shown in Table~\ref{tab:general_vqa_comparison}, MiniCPM-V 4.5 demonstrates strong performance across a wide range of vision-language capabilities. 

\textbf{Comprehensive Capability.} MiniCPM-V 4.5 achieves an average score of 77.0 on OpenCompass, a comprehensive evaluation of 8 popular benchmarks. With only 8B parameters, it surpasses widely used proprietary models like GPT-4o-latest and strong open-source models like Qwen2.5-VL 72B for vision-language capabilities.

\textbf{Video Understanding.} The model achieves strong performance on high frame rate and fine-grained action dynamics video benchmarks such as MotionBench and FlavorBench. It also shows competitive performance on long video understanding benchmarks such as VideoMME, LVBench, MLVLU, LongVideoBench, etc. 

\textbf{OCR and Document Analysis.} MiniCPM-V 4.5 achieves leading performance on OCRBench, surpassing proprietary models such as GPT-4o-latest. It also achieves state-of-the-art performance for PDF document parsing capability on OmniDocBench among general MLLMs. 

\textbf{Hallucination Reduction.} The model shows a significant reduction in visual hallucinations, outperforming other models on ObjectHalBench and MMHal-Bench, since the RLAIF-V training stage specifically enhances the level of trustworthiness.

\begin{table}[t]
\centering

\begin{subfigure}[t]{0.48\textwidth}
\centering
\resizebox{\textwidth}{!}{
\begin{tabular}{lccc}
\toprule
\textbf{Model} & \textbf{Size} & \textbf{Avg Score} $\uparrow$ & \textbf{Time} $\downarrow$ \\
\midrule
GLM-4.1V-9B-thinking & 10.3B & 76.6 & 17.5h \\
MiMo-VL-7B-RL        & 8.3B  & 76.4 & 11.0h   \\
MiniCPM-V 4.5        & 8.7B  & \textbf{77.0} & \textbf{7.5h}  \\
\bottomrule
\end{tabular}}
\caption{OpenCompass results of thinking models }
\label{tab:opencompass-efficiency}
\end{subfigure}
\hfill
\begin{subfigure}[t]{0.51\textwidth}
\centering
\resizebox{\textwidth}{!}{
\begin{tabular}{lcccc}
\toprule
\textbf{Model} & \textbf{Size} & \textbf{Score} $\uparrow$ & \textbf{Time} $\downarrow$ & \textbf{Mem} $\downarrow$ \\
\midrule
Qwen2.5-VL-7B & 8.3B  & 71.6 & 3.00h    & 60G \\
GLM-4.1V-9B-thinking   & 10.3B & \textbf{73.6} & 2.63h & 32G \\
MiniCPM-V 4.5          & 8.7B  & 73.5 & \textbf{0.26h} & \textbf{28G} \\
\bottomrule
\end{tabular}}
\caption{Video-MME results }
\label{tab:videomme-efficiency}
\end{subfigure}
\caption{Inference efficiency on 8 A100 GPUs. Best results are marked in \textbf{bold}. }
\label{tab:inference-efficiency}
\end{table}

\subsection{Inference Efficiency}

We evaluated the inference efficiency of MiniCPM-V 4.5 in a standard configuration of 8 A100 GPUs on both image understanding and video understanding tasks. As detailed in Table~\ref{tab:inference-efficiency}, our model achieves competitive or superior performance while significantly reducing inference time and GPU memory consumption compared to other leading models. On OpenCompass, MiniCPM-V 4.5 not only achieves the highest average score among models under 30B, but also finishes the evaluation using 42.9\% of the time of GLM-4.1V. This efficiency is enabled by the model's flexible short and long reasoning modes. On VideoMME, the model demonstrates remarkable efficiency gains. With a strong performance of 73.6, it also reduces the inference time by nearly 10$\times$ (from 2.63h to 0.26h) and uses the least memory of 28G. This improvement is primarily due to the efficient 3D-Resampler, which compresses videos jointly considering spatial and temporal dimensions.

\subsection{Ablations}

We ablate key design choices of MiniCPM-V 4.5 in this section. 

\begin{floatingfigure}[r]{0.48\textwidth}
  \centering
  \resizebox{0.48\textwidth}{!}{%
    \begin{tabular}{lcc}
      \toprule
      \textbf{Method} & \textbf{OpenCompass} & \textbf{Training Tokens}\\
      \midrule
      Short reasoning only & 76.0 & \textbf{1.6B}\\
      Long reasoning only  & 77.0 & 4.4B\\
      Hybrid               & \textbf{77.1} & 3.1B\\
      \bottomrule
    \end{tabular}
  }
  \captionof{table}{Ablation of hybrid reinforcement learning. 
                    We report training token cost and performance on OpenCompass.}
  \label{tab:ablation_hybrid}
\end{floatingfigure}

\textbf{Hybrid reasoning reinforcement learning helps improve overall performance and efficiency.}
We evaluate the hybrid RL strategy that mixes samples from both long and short reasoning modes during training. As shown in Table~\ref{tab:ablation_hybrid}, we observe that the hybrid strategy achieves the best multimodal understanding performance, demonstrating it effectively incentivizes strong long reasoning capability with only half of the long reasoning samples during training. Moreover, the hybrid strategy consumes only 70.5\% of the training token costs of the long reasoning only setting to achieve better performance. We hypothesize that this is because both modes share foundational perceptual and cognitive skills. The analytical depth cultivated by long reasoning appears to bolster the short reasoning, while the efficiency and directness learned from the short reasoning refine the long reasoning process.

\begin{figure}[tbp]
    \centering
    \begin{subfigure}[b]{0.325\textwidth}
        \centering
        \includegraphics[width=\textwidth]{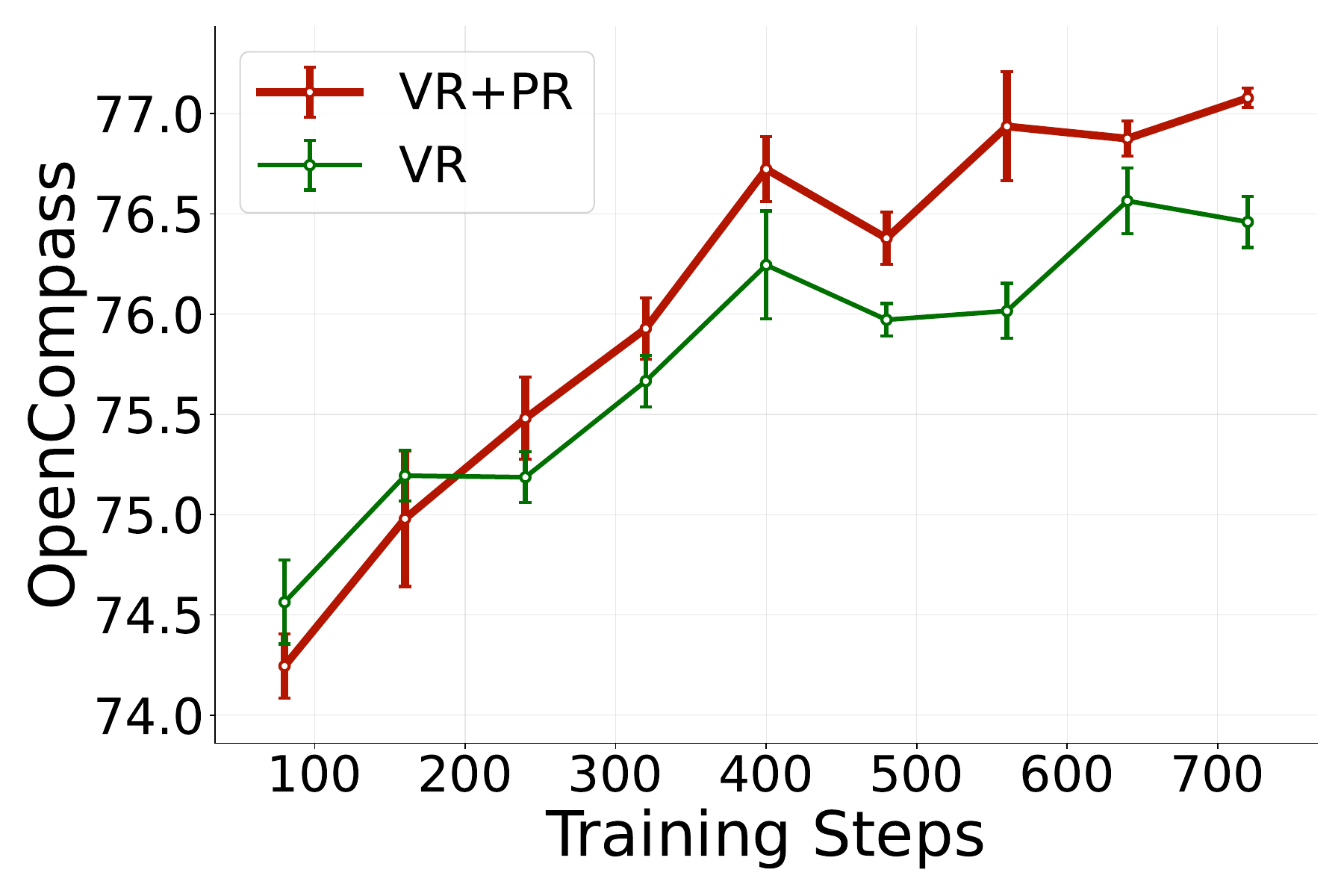}
        \vspace{-3mm}
        \captionsetup{justification=centering, singlelinecheck=false}
        \caption{OpenCompass}
    \end{subfigure}
    \hfill
    \begin{subfigure}[b]{0.325\textwidth}
        \centering
        \includegraphics[width=\textwidth]{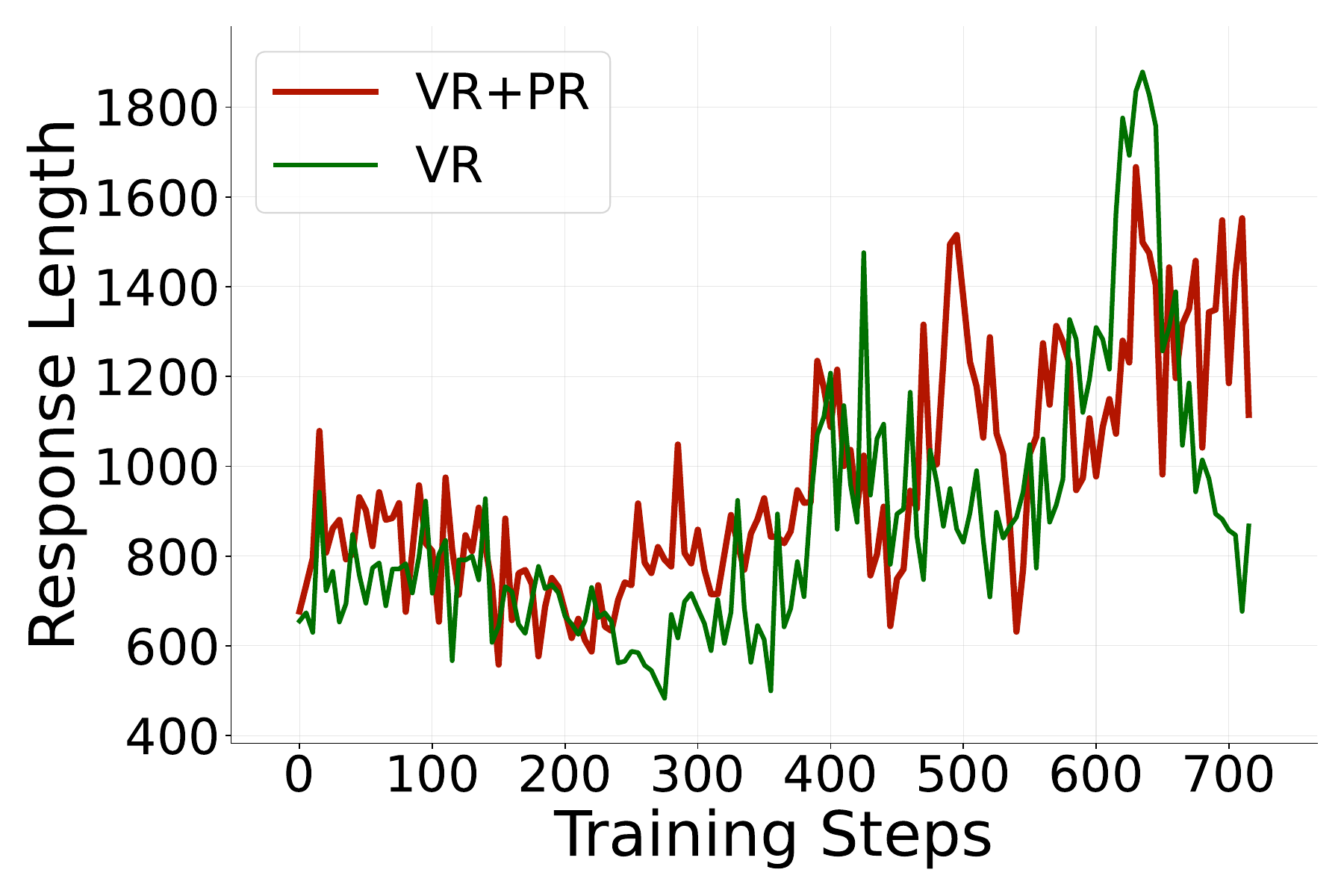}
        \vspace{-3mm}
        \captionsetup{justification=centering, singlelinecheck=false}
        \caption{Response Length}
    \end{subfigure}
    \hfill
    \begin{subfigure}[b]{0.325\textwidth}
        \centering
        \includegraphics[width=\textwidth]{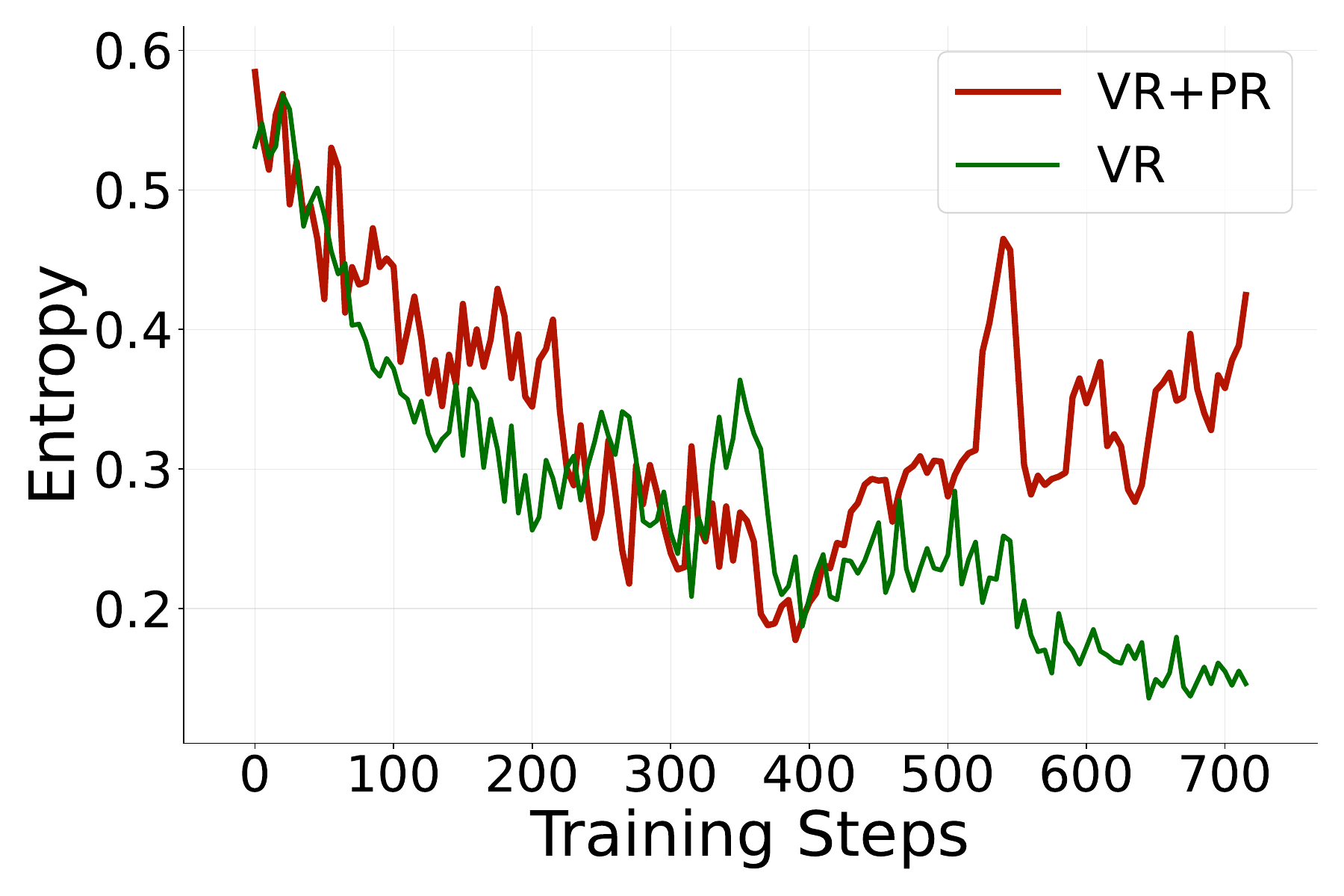}
        \vspace{-3mm}  
        \captionsetup{justification=centering, singlelinecheck=false}
        \caption{Entropy}
    \end{subfigure}
    
    \caption{Performance ablation of adding probability-based reward. We report OpenCompass scores, response length, and entropy on different training steps. }
    \label{fig:vrpr}
\end{figure}

\textbf{Probability-based reward complements rule-verification reward.} In addition to rule-based reward for easy-to-verify responses, MiniCPM-V 4.5 further incorporates the probability-based reward from RLPR~\citep{yu2025rlprextrapolatingrlvrgeneral} for general domain response verification.
As shown in Figure \ref{fig:vrpr}, combining both rule-based and probability-based signals (VR + PR) consistently and substantially outperforms the rule-only approach, while also yielding stable training patterns with respect to response length and entropy. This confirms that probability-based reward provides a meaningful learning signal for the general reasoning data that rules struggle with, effectively complementing the small subset of simple data suitable for rule verification. The effectiveness becomes particularly evident as the number of training steps scales, where the robust reward signals across the full spectrum of multimodal scenarios provide essential training guidance that pure rule-based verification cannot deliver.

\begin{table}[t]
\begin{minipage}{0.48\textwidth}
\centering
\resizebox{\linewidth}{!}{
\begin{tabular}{lccc}
\toprule
\textbf{Method} & \textbf{MMMU} & \textbf{AI2D} & \textbf{OCRBench} \\
\midrule
External Parser    & 49.0 & 74.9 & 576  \\
Unified Learning & \textbf{51.4} & \textbf{76.5}  & \textbf{617}  \\
\bottomrule
\end{tabular}
}
\vspace{3mm}
\captionof{table}{Ablation of unified learning paradigm for document knowledge and text recognition. We report results on knowledge-intensive, document understanding, and text recognition benchmarks.}
\label{tab:ablation_unified_doc_ocr}
\end{minipage}%
\hfill
\begin{minipage}{0.48\textwidth}
\centering
\vspace*{-3.6mm}
\resizebox{\linewidth}{!}{
\begin{tabular}{lccc}
\toprule
\textbf{Method} & \textbf{w/ sub} & \textbf{w/o sub} & \textbf{tokens/frame} \\
\midrule
2D-Resampler & 65.5 & 71.5 & 64.0 \\
3D-Resampler & \textbf{67.3} & \textbf{72.5} & \textbf{21.3}   \\
\bottomrule
\end{tabular}
}
\vspace{3mm}
\captionof{table}{Ablation of the 3D-Resampler. 
We report scores on VideoMME. w/ sub: using subtitles during evaluation; w/o sub: remove subtitles during evaluation}
\label{tab:ablation_3d_resampler}
\end{minipage}
\vspace{-3mm}
\end{table}

\textbf{Unified learning of document knowledge and text recognition improves both capabilities.} 
We run an ablation experiment for the proposed unified learning paradigm. Following the three stages pre-training process in §~\ref{sec:pretrain}, we train the model on 1M high-quality samples, 20\% of which are knowledge-intensive documents.
Then we conduct a comparison against the baseline method after the same SFT pipeline. 
As shown in Table~\ref{tab:ablation_unified_doc_ocr}, the unified approach outperforms the baseline on both knowledge-intensive evaluations and text-recognition tasks. 
These gains indicate that learning directly from document images mitigates the noise introduced by fragile external parsers.

\textbf{3D-Resampler enables higher performance with lower token cost.} We ablate the 3D-Resampler to verify its effectiveness. To ensure a fair comparison against the 2D baseline, we fine-tuned the model ckpt after the general SFT stage for 300 steps, isolating the resampler architecture as the only variable.
As demonstrated in Table~\ref{tab:ablation_3d_resampler}, our 3D-Resampler achieves stronger performance, while using only one-third of the visual tokens per frame required by the 2D baseline.

\section{Conclusion}

We introduce MiniCPM-V 4.5, an MLLM designed with high efficiency at both training and inference time via architecture, data, and training recipe. With a unified 3D-Resampler, it achieves strong performance on high frame rate and long video understanding with superior encoding efficiency. Furthermore, the unified learning paradigm for document knowledge and text recognition allows the model to directly learn from document images. This approach bypasses fragile parsers and significantly reduces the data engineering complexity. Finally, the hybrid post-training strategy improves both training and inference efficiency while also facilitating generalization between short and long reasoning modes. Overall, MiniCPM-V 4.5 demonstrates a promising path toward addressing the efficiency bottlenecks in MLLM development.

{
\bibliographystyle{unsrt}
}
\bibliography{neurips_2024}

\appendix

\newpage
\section{Implementation Details}

Pre-training follows a WSD schedule~\cite{hu2024minicpm} with a fixed learning rate of $5\times10^{-5}$ in the stable phase, decaying to $1\times10^{-5}$. SFT applies cosine decay from $1\times10^{-5}$ to $1\times10^{-6}$. The Long-CoT and 3D-Resampler stage continues from the SFT checkpoint, warming up to $5\times10^{-6}$ and decaying to $1\times10^{-6}$.

For the RL stage, we adopt GRPO~\citep{shao2024deepseekmath} without entropy loss or KL penalty. Each batch consists of $128$ prompts with $8$ responses each, and a max response length of $8192$ tokens to support detailed reasoning. Rollouts use a temperature of $1.0$, with $50\%$ of prompts assigned to long reasoning mode. We use a fixed learning rate of $1\times10^{-6}$ throughout RL. In the RLAIF-V~\cite{yu2024rlaifvopensourceaifeedback} stage, we use a global batch size of $256$, learning rate of $1\times10^{-6}$, and $\beta=0.1$ for $400$ steps. 

\section{Qualitative Cases}

\subsection{Comprehensive Instruction Following}

\begin{figure}[h]
    \centering
    \includegraphics[width=\linewidth]{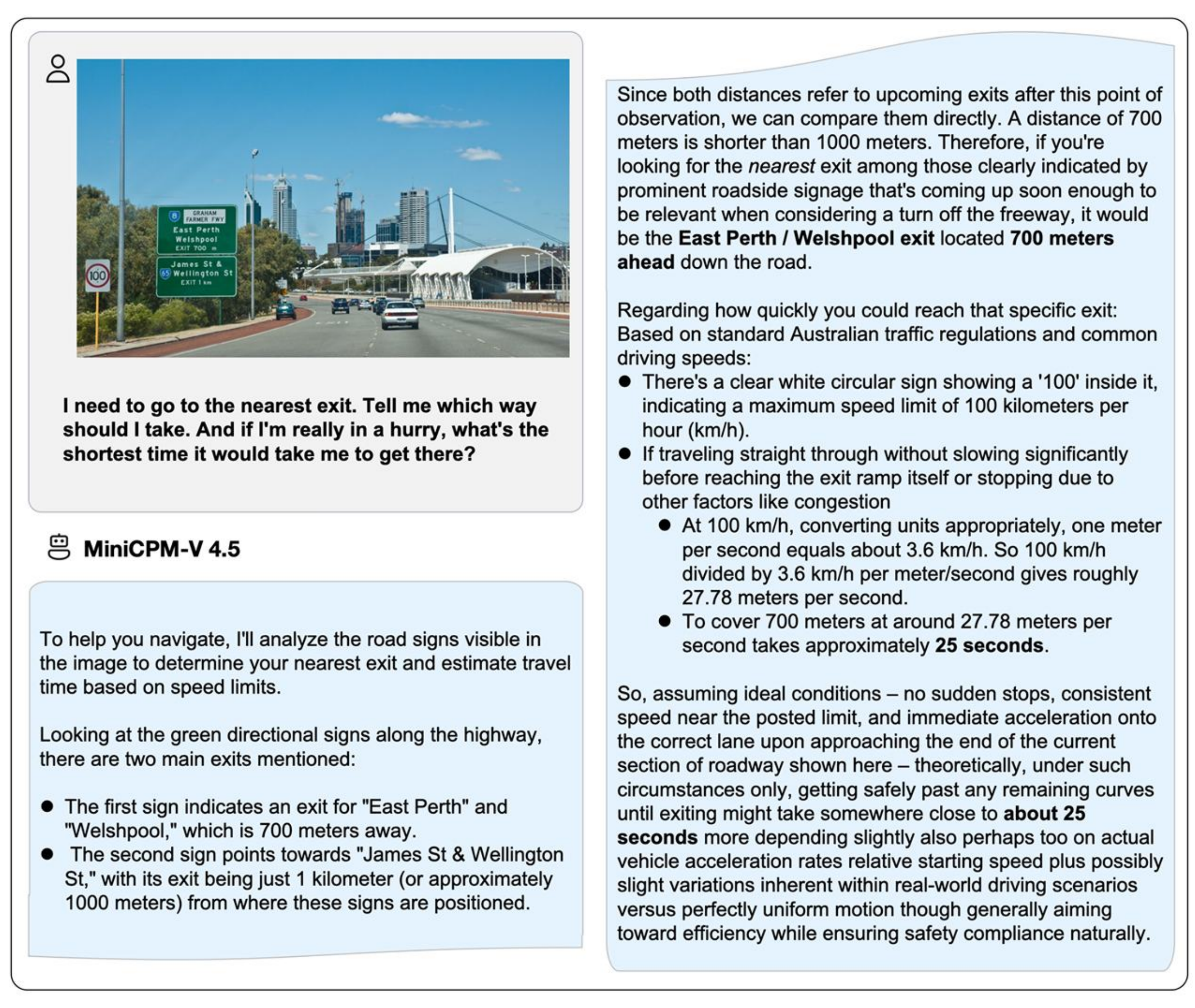}
    \caption{A case of comprehensive real-world reasoning. }
\end{figure}

\begin{figure}[h]
    \centering
    \includegraphics[width=\linewidth]{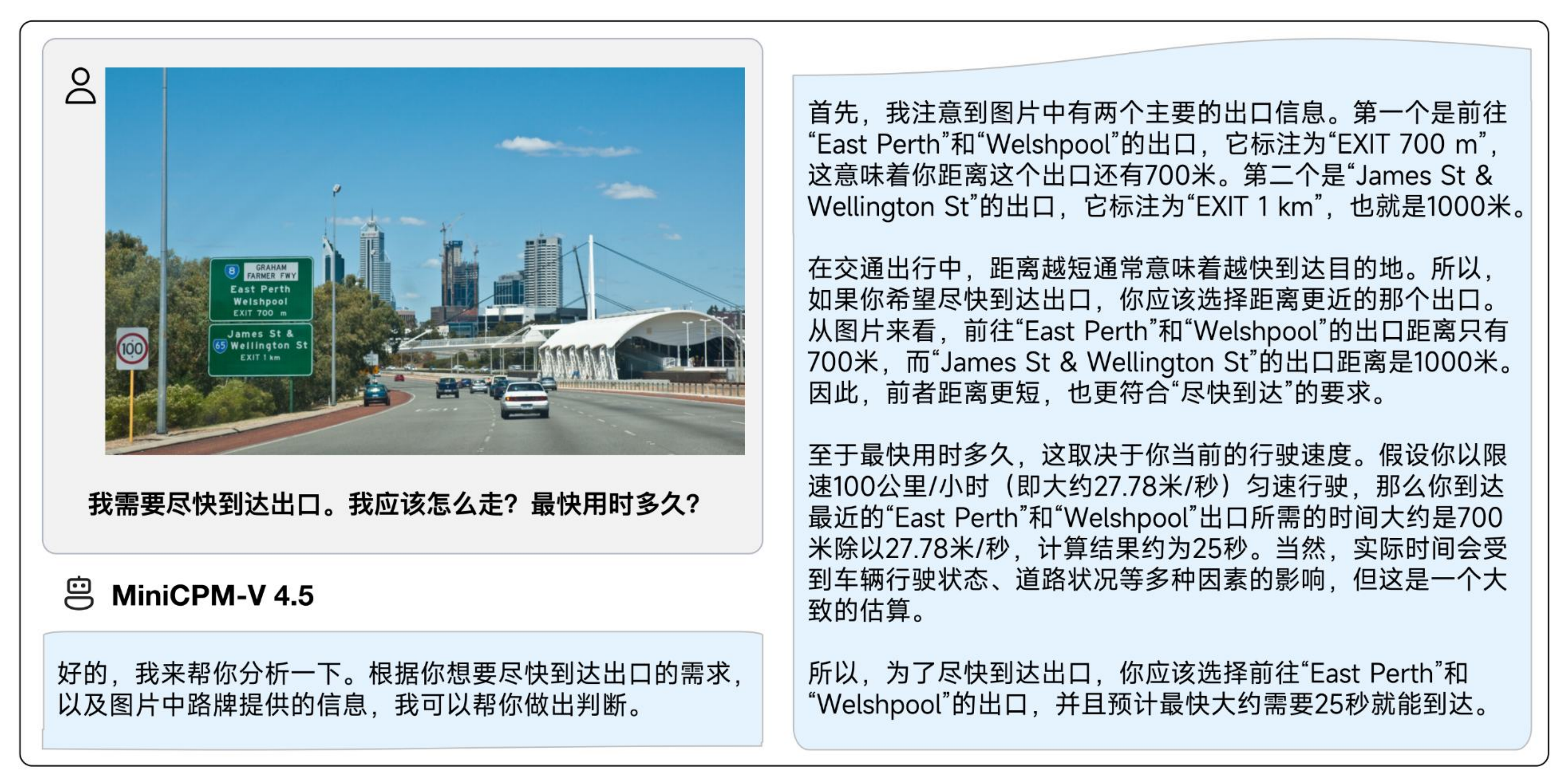}
    \caption{A case of comprehensive real-world reasoning in Chinese. }
\end{figure}

\begin{figure}[h]
    \centering
    \includegraphics[width=\linewidth]{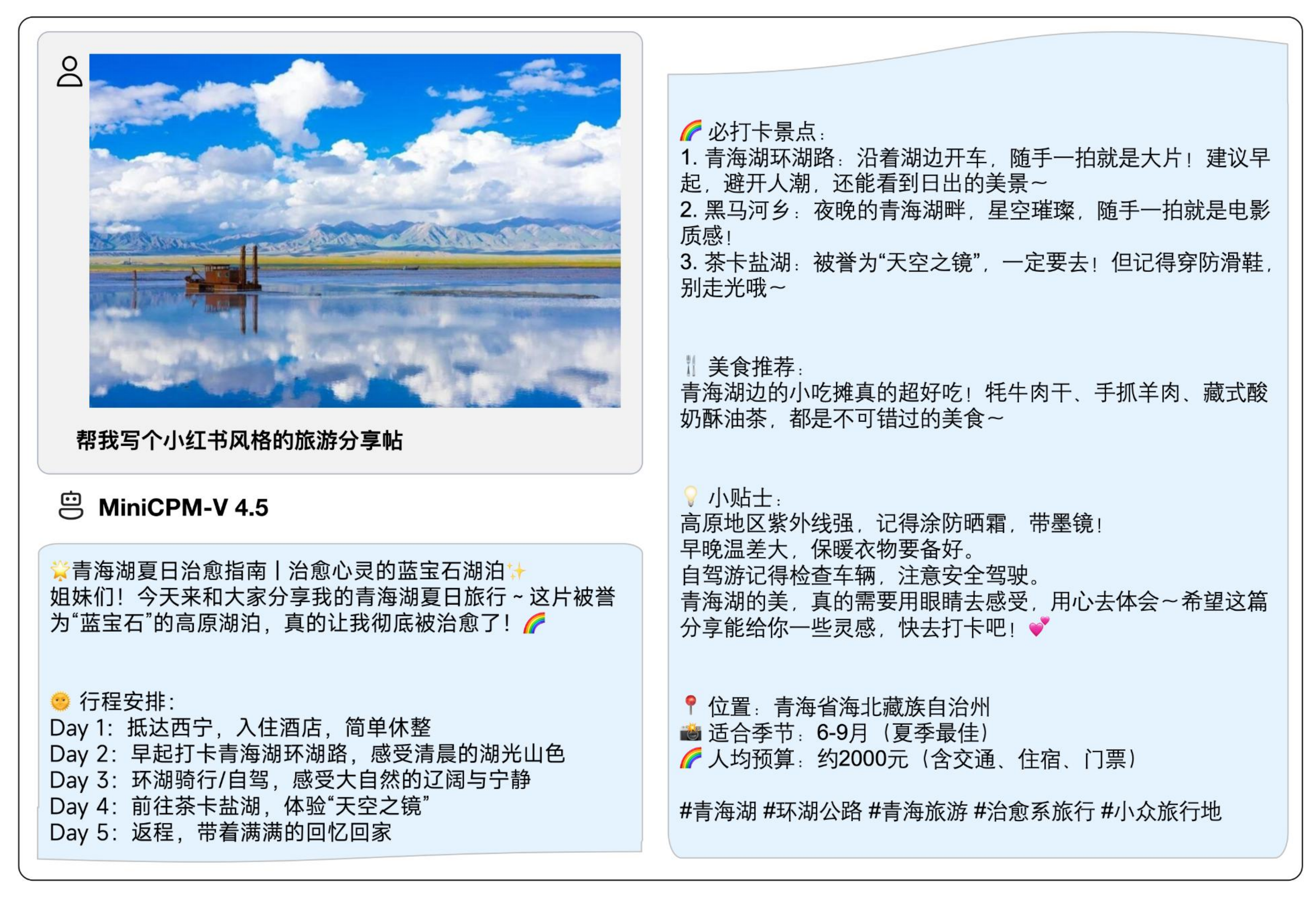}
    \caption{A case of creative writing in Chinese. }
\end{figure}
\clearpage
\subsection{World Knowledge}

\begin{figure}[h]
    \centering
    \includegraphics[width=\linewidth]{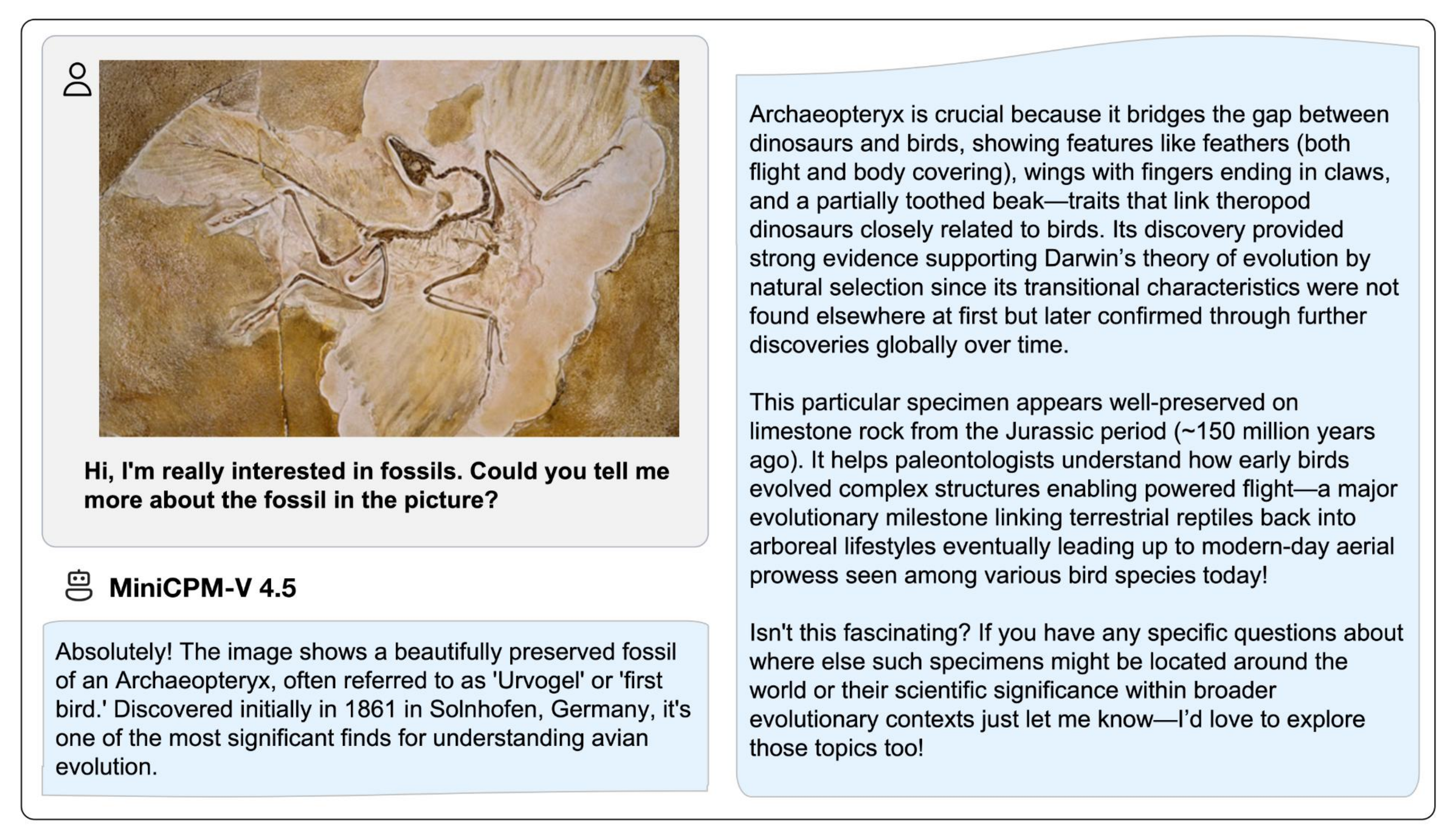}
    \caption{A case of world knowledge understanding. }
\end{figure}

\begin{figure}[h]
    \centering
    \includegraphics[width=\linewidth]{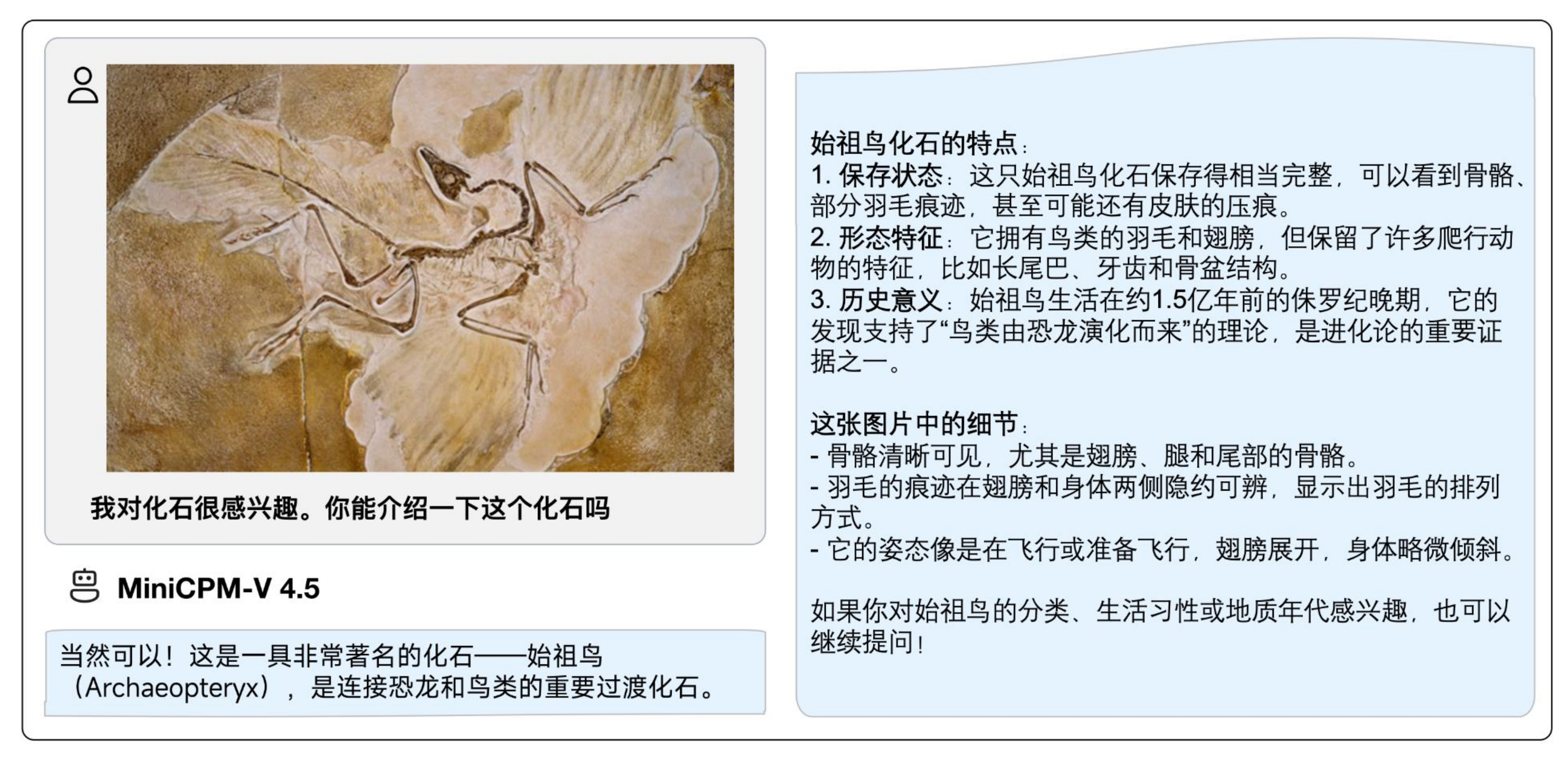}
    \caption{A case of world knowledge understanding in Chinese. }
\end{figure}

\clearpage
\subsection{OCR}

\begin{figure}[h]
    \centering
    \includegraphics[width=\linewidth]{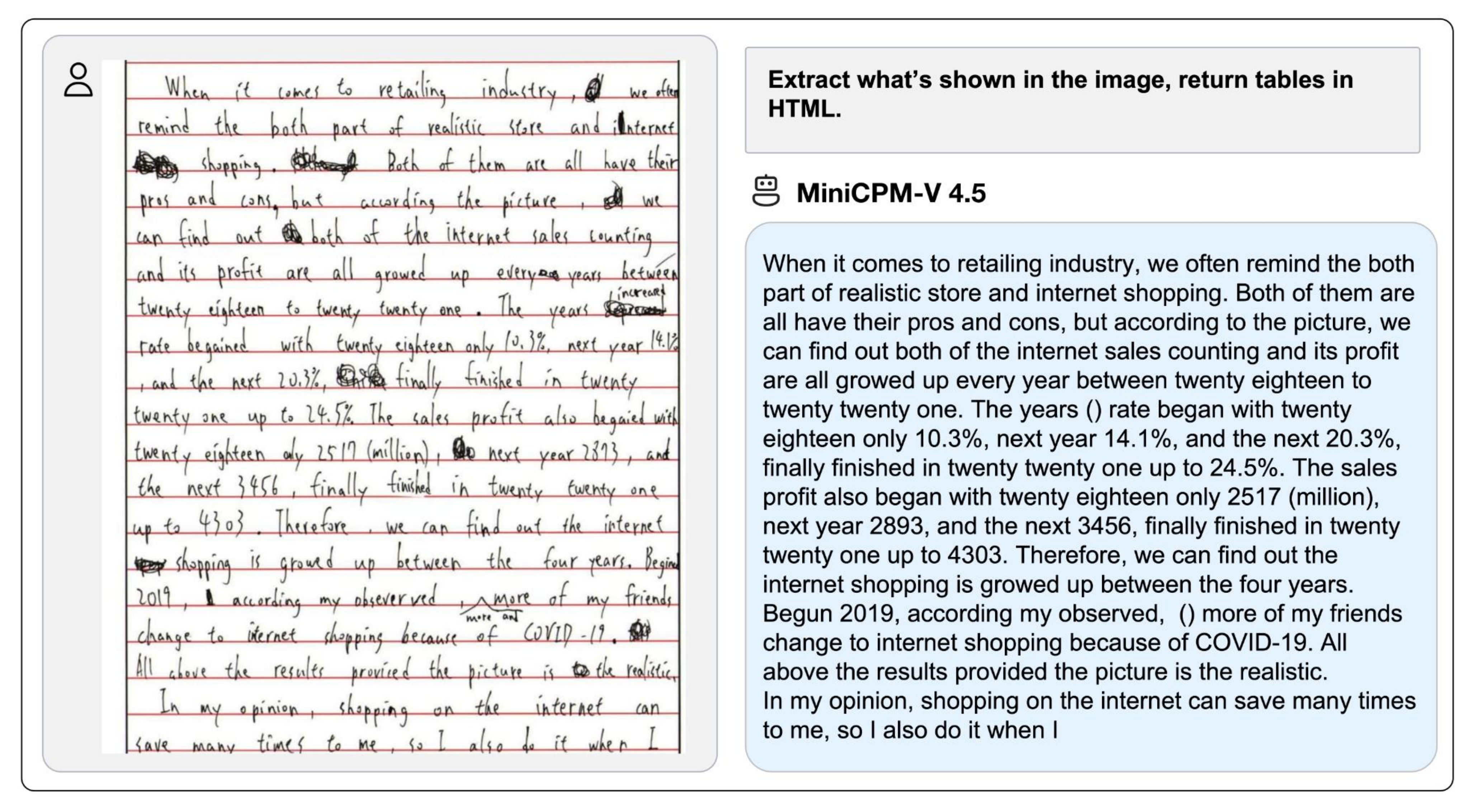}
    \caption{A case of handwritten text recognition. }
\end{figure}

\begin{figure}[h]
    \centering
    \includegraphics[width=\linewidth]{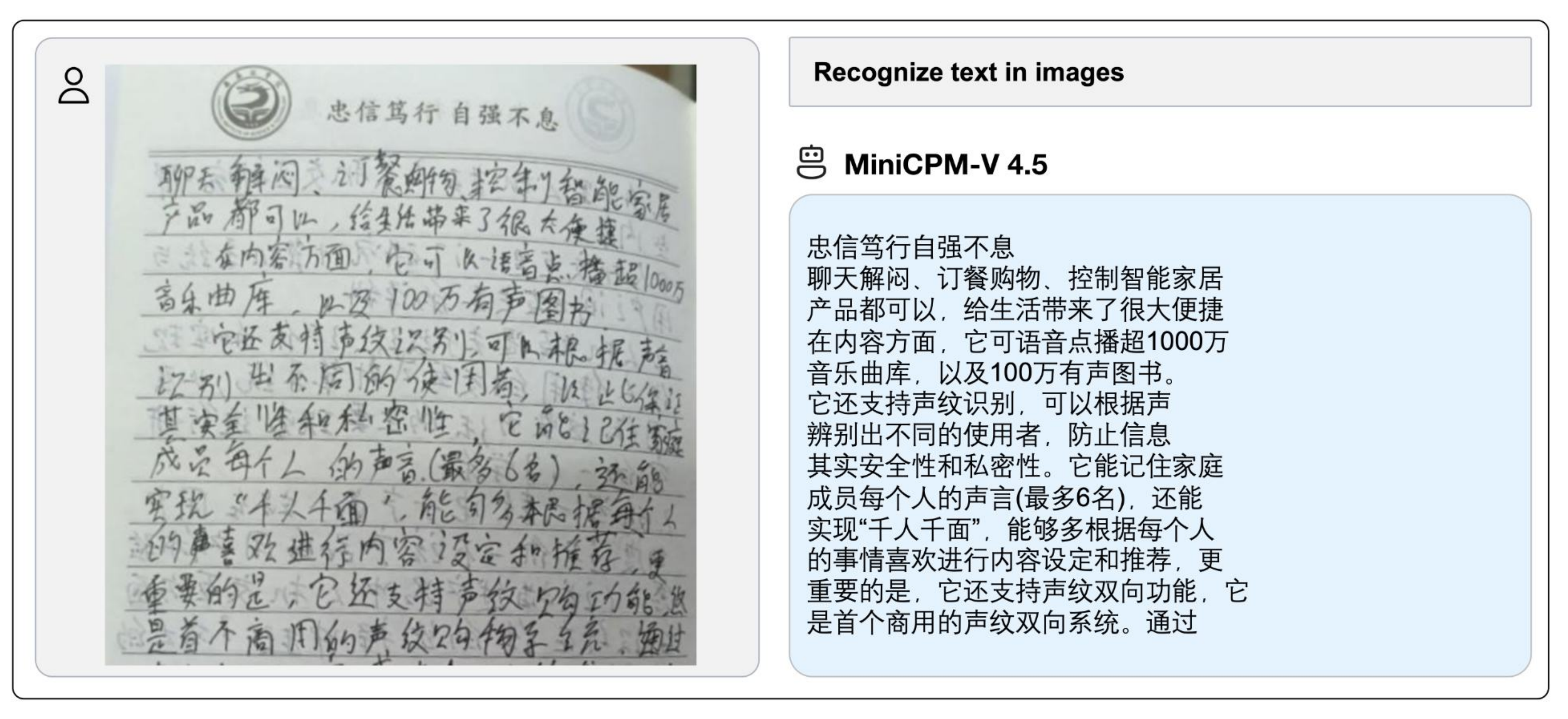}
    \caption{A case of handwritten text recognition in Chinese. }
\end{figure}

\begin{figure}[h]
    \centering
    \includegraphics[width=\linewidth]{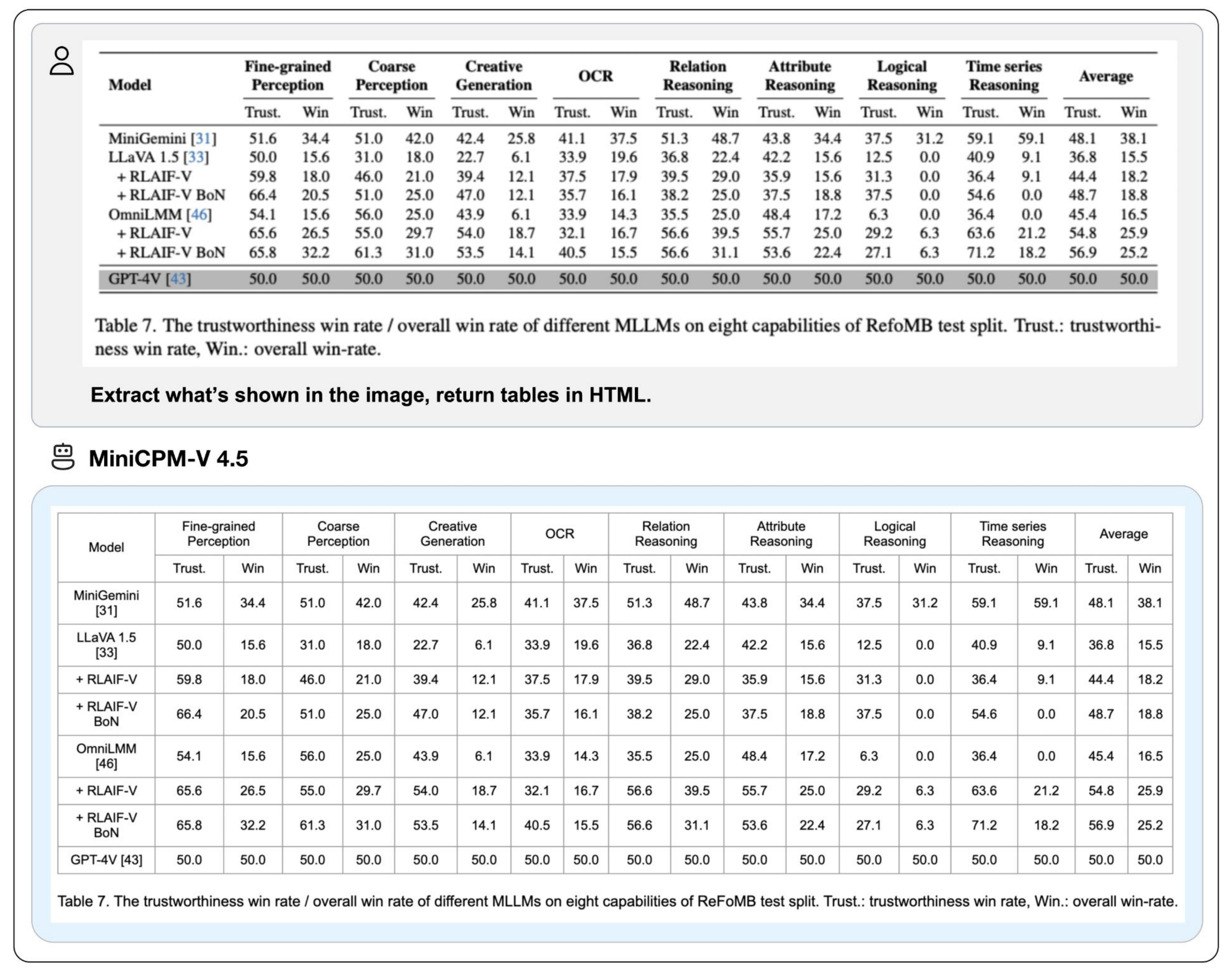}
    \caption{A case of table content extraction. }
\end{figure}

\newpage
\subsection{Problem Solving}

\begin{figure}[h]
    \centering
    \includegraphics[width=\linewidth]{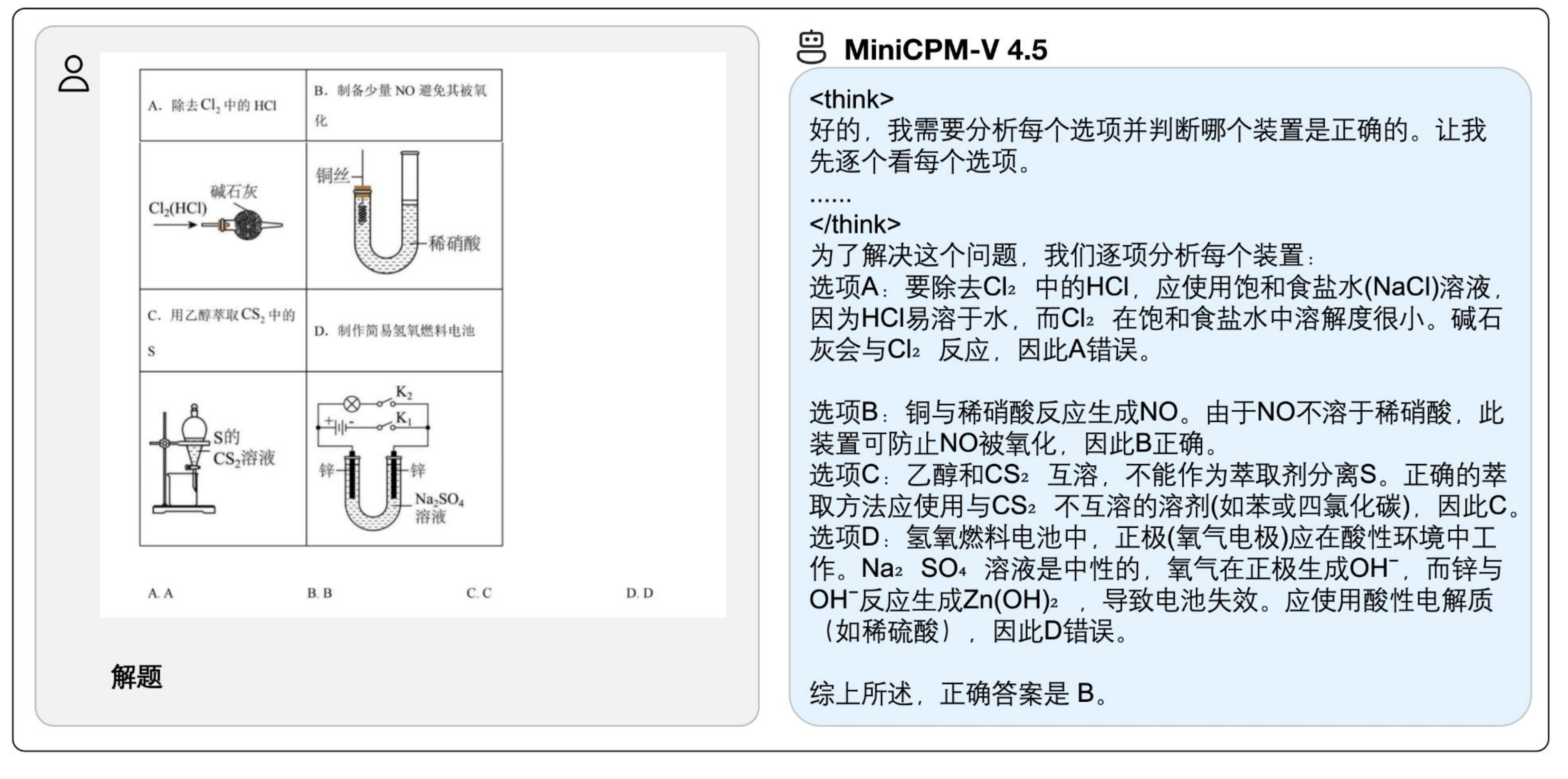}
    \caption{A case of chemistry problem solving in Chinese. }
\end{figure}

\begin{figure}[h]
    \centering
    \includegraphics[width=\linewidth]{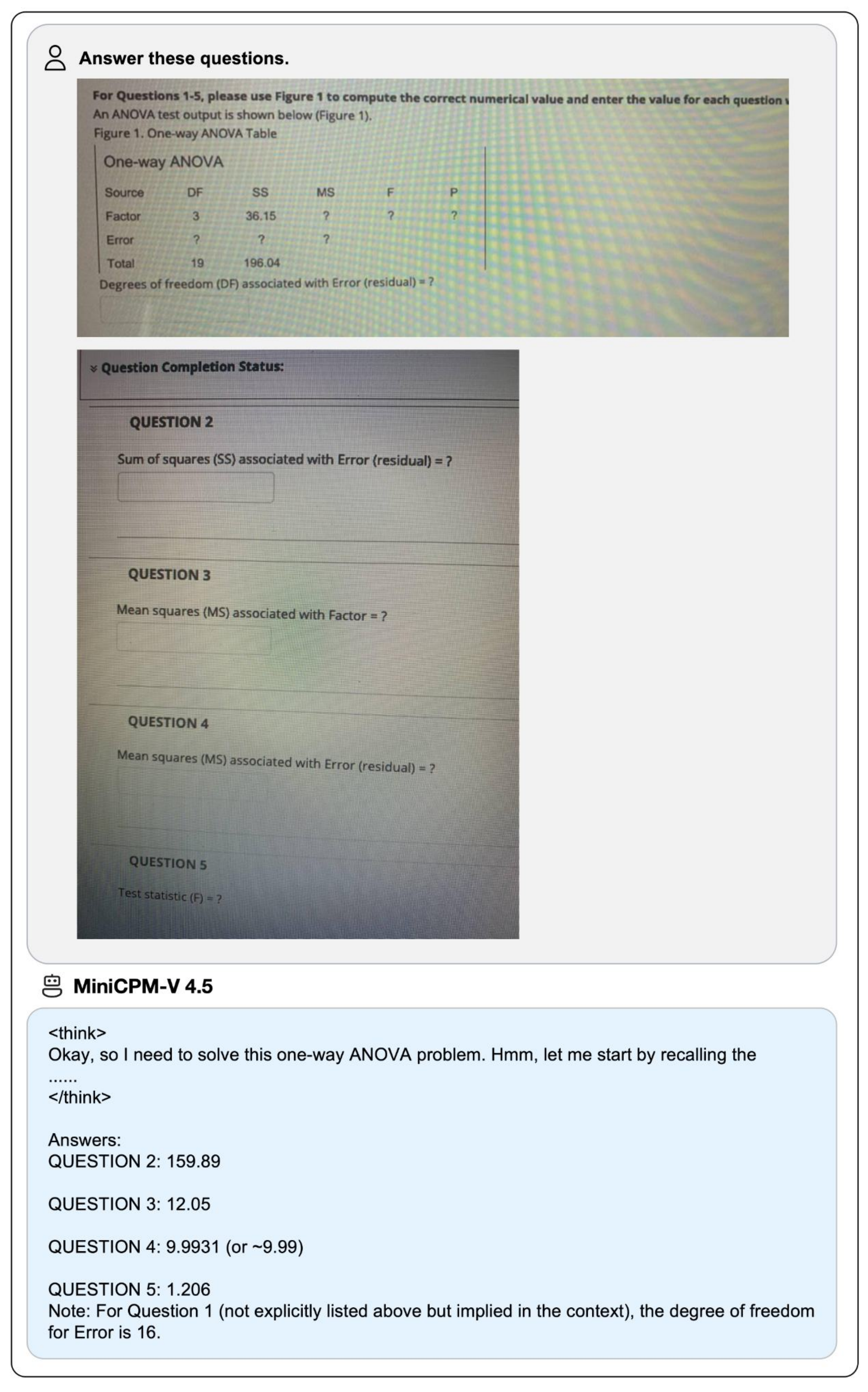}
    \caption{A case of multi-image statistical problem solving. }
\end{figure}

\end{document}